\documentclass[10pt,twocolumn,letterpaper]{article}

\usepackage{cvpr}
\usepackage{times}
\usepackage{epsfig}
\usepackage{graphicx}
\usepackage{amsmath}
\usepackage{amssymb}
\usepackage{mathdots}
\usepackage{mdwmath}
\usepackage{mdwtab}
\usepackage{bm}
\usepackage{myothers}

\usepackage[pagebackref=true,breaklinks=true,letterpaper=true,colorlinks,bookmarks=false]{hyperref}

\cvprfinalcopy 

\def\cvprPaperID{2129} 

\ifcvprfinal\fi
\begin{document}

\title{High-speed Tracking with Multi-kernel Correlation Filters}


\author{Ming Tang$^{1,2}$\thanks{The corresponding author (tangm@nlpr.ia.ac.cn). This work was supported by Natural Science Foundation of China under
Grants 61375035 and 61772527. The code is available at http://www.nlpr.ia.ac.cn/mtang/ Publications.htm.}, Bin Yu$^{1,2}$, Fan Zhang$^3$, and Jinqiao Wang$^{1,2}$\\
$^1$University of Chinese Academy of Sciences, Beijing, China\\
$^2$National Lab of Pattern Recognition, Institute of Automation, CAS, Beijing 100190, China\\
$^3$School of Info. \& Comm. Eng., Beijing University of Posts and Telecommunications
}


\maketitle

\begin{abstract}
Correlation filter (CF) based trackers are currently ranked top in terms of their performances. Nevertheless, only some of them, such as KCF~\cite{henriques15} and MKCF~\cite{tangm15}, are able to exploit the powerful discriminability of non-linear kernels. Although MKCF achieves more powerful discriminability than KCF through introducing multi-kernel learning (MKL) into KCF, its improvement over KCF is quite limited and its computational burden increases significantly in comparison with KCF. In this paper, we will introduce the MKL into KCF in a different way than MKCF. We reformulate the MKL version of CF objective function with its upper bound, alleviating the negative mutual interference of different kernels significantly. Our novel MKCF tracker, MKCFup, outperforms KCF and MKCF with large margins and can still work at very high fps. Extensive experiments on public data sets show that our method is superior to state-of-the-art algorithms for target objects of small move at very high speed.
\end{abstract}
\vspace{-3mm}
\section{Introduction}
\label{sec:introduction}


Visual object tracking is one of the most challenging problems in computer vision~\cite{tangm2012,hongz14,kwon14,nam14,leed14,mac15a,liut15,mac15b,huay15,zhangt16,yunsd17,hanby17,valmadre17,choijw17,rozumnyid17}. To adapt to unpredictable variations of object appearance and background during tracking, the tracker could select a single strong feature that is robust to any variation. However, this strategy has been known to be difficult~\cite{varma2007,gehler2009}, especially for a model-free tracking task in which no prior knowledge about the target object is known except for the initial frame. Therefore, designing an effective and efficient scheme to combine several complementary features for tracking is a reasonable alternative~\cite{wuyi2011,yangf2014,lanx2014,dane14a,bertinetto16,wanglj16,zhangt17a,zhangl17}.


\begin{figure}
	\centering
        \includegraphics[width=0.49\textwidth]{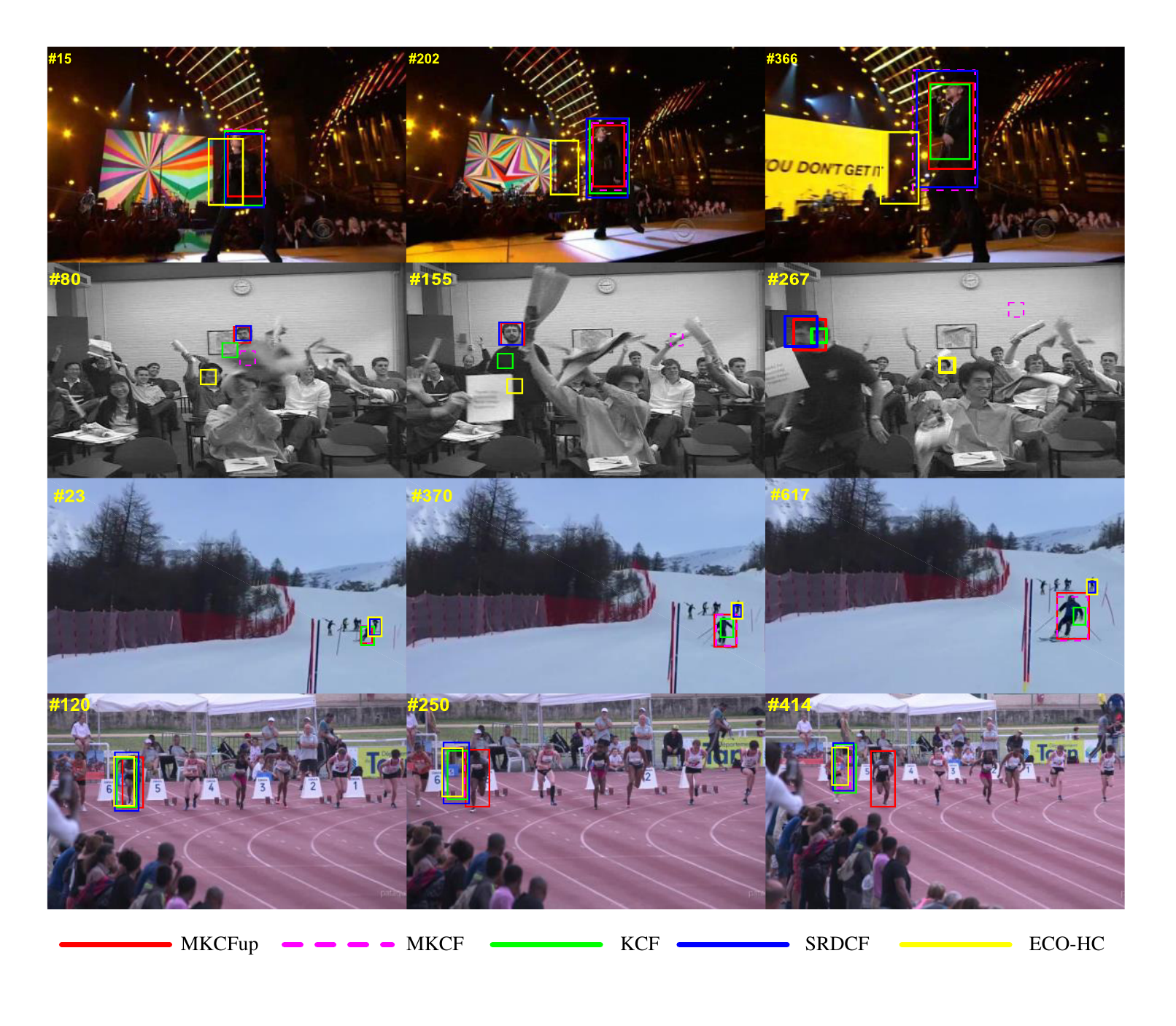}
\caption{Qualitative comparison of our novel multi-kernel correlation filters tracker, MKCFup, with state-of-the-art trackers, KCF~\cite{henriques15}, MKCF~\cite{tangm15}, SRDCF~\cite{dane15a}, and ECO\_HC~\cite{dane17b} on challenging sequences, singer2 and freeman4 of OTB2013~\cite{wu13} and ski\_long and running\_100\_m\_2 of NfS~\cite{galoo17b}.}
\label{fig:intro-qualcomp}
\vspace{-4mm}
\end{figure}

Since 2010, correlation filter based trackers (CF trackers) have been being proposed and almost dominated the tracking domain in recent years~\cite{bolme10,henriques2012,dane14a,dane14b,henriques15,dane16a,dane16b,choijw16,cuiz16,qiyk16,dane17a,mueller17,luke17}. Bolme~\emph{et al.}~\cite{bolme10} reignited the interests in correlation filters in the vision community by proposing a CF tracker, called minimum output sum of squared error (MOSSE), with classical signal processing techniques. MOSSE used a base image patch and several virtual ones to train the correlation filter directly in the Fourier domain, achieving top accuracy and fps then. Later, the expression of MOSSE in the spatial domain turned out to be the ridge regression~\cite{rifkin03} with a linear kernel~\cite{henriques2012}. Therefore, in order to exploit the powerful discriminability of non-linear kernels, Henriques~\etal~\cite{henriques2012,henriques15} utilized the circulant structure produced by a base sample to propose an efficient kernelized correlation filter based tracker (KCF). Danelljan~\emph{et al.}~\cite{dane14a} extended the KCF with the historically weighted objective function and low-dimensional adaptive color channels. To adaptively employ complementary features in KCF, Tang and Feng~\cite{tangm15} derived a multi-kernel learning (MKL)~\cite{rako08} based correlation filter (MKCF) which is able to take advantage of the invariance-discriminative power spectrums of various features~\cite{varma2007} to improve the location performance. By introducing a mask on the samples into the loss item of correlation filter formulation, Galoogani \etal~\cite{galoo15} proposed the correlation filter with limited boundaries (CFLB) to address the boundary effect~\cite{kumar05}. And Danelljan \etal~\cite{dane15a} introduced a smooth spatial regularization factor within the regularizer to restrain the boundary effect. In~\cite{dane17b}, Danelljan \etal employed the dimensionality reduction, linear weighting of features, and sample clustering to further improve the SRDCF proposed in~\cite{dane15a} in both location accuracy and fps. 

Up till now, there are at least two principal lines to improve MOSSE and KCF. The first one is to weight the filter or samples with a mask in MOSSE or the KCF of linear kernel, alleviating the negative boundary effect greatly and improving the location performance remarkably. However, the trackers on this line, such as CFLB, SRDCF, C-COT~\cite{dane16b}, and ECO\_HC~\cite{dane17b}, are unable to employ powerful non-linear kernels. And the other line is to improve the objective function of KCF, such as designing more complicated objective functions~\cite{bibi16}, or introducing the MKL into KCF to adaptively exploit multiple (non-linear) kernels. Although MKCF, the MKL version of KCF, is more discriminative than KCF, its improvement over KCF is quite limited because different kernels of MKCF may restrict each other in training and updating. And unfortunately, the computational cost of MKCF increases significantly in comparison to KCF. Specifically, the MKCF's improvement over KCF on AUC is only about $2\%\sim3\%$, while its fps drops dramatically from averagely about 300 of KCF to 30. It is noticed that such an improvement of introducing MKL into KCF is similar to that of introducing MKL into single kernel binary classifier~\cite{varma2007}, where the improvement of MKL version is about $2\%$.

In this paper, we will introduce the MKL into KCF in a different way than~\cite{tangm15} to adaptively exploit multiple complementary features and non-linear kernels more effectively than in MKCF. We reformulate the MKL version of CF objective function with its upper bound, alleviating the negative mutual interference of complementary features significantly while keeping very large fps. In fact, our novel MKCF tracker, \ie, MKCFup, outperforms KCF and the KCF with scaling on AUC about $16\%$ and $7\%$, respectively, at about 150 fps. A qualitative comparison shown in Fig.~\ref{fig:intro-qualcomp} indicates that our novel tracker, MKCFup, outperforms other state-of-the-art trackers in challenging sequences singer2 and freeman4 of OTB2013~\cite{wu13} and ski\_long and running\_100\_m\_2 of NfS~\cite{galoo17b}. 

The remainder of this paper is organized as follows. In Sec.\ref{sec:relatedwork}, we briefly overview the related work. Sec.\ref{sec:mkcfup} first simplifies the solution of MKCF, then analyzes its shortcoming, and finally derives a novel multi-kernel correlation filter with the upper bound of objective function. Sec.\ref{sec:details} provides some necessary implementation details. Experimental results and comparison with state-of-the-art approaches are presented in Sec.\ref{sec:exper}. Sec.\ref{sec:conclusion} summarizes our work.

\section{Related Work}
\label{sec:relatedwork}
Multi-kernel learning (MKL) aims at simultaneously learning a kernel and the associated predictor in supervised learning settings. Rakotomamonjy~\emph{et al.}~\cite{rako08} proposed an efficient algorithm, named SimpleMKL, for solving the MKL problem through reduced gradient descent in a primal formulation. Varma and Ray~\cite{varma2007} extended the MKL formulation in~\cite{rako08} by introducing an additional constraint on combinational coefficients and applied it to object classification. Vedaldi~\etal~\cite{vedaldi09} and Gehler and Nowozin~\cite{gehler2009} applied MKL based approaches to object detection and classification. Cortes~\etal~\cite{cortes09a} studied the problem of learning kernels of the same family with an $L_2$ regularization for ridge regression (RR)~\cite{rifkin03}. Tang and Feng~\cite{tangm15} extended the MKL formulation of~\cite{rako08} to RR, and presented a different multi-kernel RR approach. In this paper, differently from all above approaches, we derive a novel multi-kernel correlation filter through optimizing the upper bound of multi-kernel version of KCF's objective function.

In addition to the correlation filter based trackers aforementioned, generalizations of KCF to other applications have also been proposed~\cite{bodd13,galoo13,henriques13} in recent years. And Henriques~\etal~\cite{henriques14} utilized the circulant structure of Gram matrix to speed up the training of pose detectors in the Fourier domain. It is noted that all these approaches are unable to employ multiple kernels or non-linear kernels simultaneously. In this paper, we propose a novel multi-kernel correlation filter which is able to fully take advantage of invariance-discriminative power spectrums of various features at really high speed.

\section{Multi-kernel Correlation Filters with Upper Bound}
\label{sec:mkcfup}
In this section, we will first review the multi-kernel correlation filter (MKCF)~\cite{tangm15}, simplify its optimization, then analyze its drawback, and finally derive a novel multi-kernel correlation filter with upper bound. Readers may refer to~\cite{rako08,gonen11} for more details on multi-kernel learning.
\subsection{Simplified Multi-kernel Correlation Filter}
\label{sec:mkcf}

The goal of a ridge regression~\cite{rifkin03} is to solve the Tikhonov regularization problem,
\begin{equation}
\label{eq:tikhonov}
	 \min_f\dfrac{1}{2}\sum_{i=0}^{l-1}(f(\mathbf{x}_i)-y_i)^{2}+\lambda_o||f||^2_k,
\end{equation}
where $l$ is the number of samples, $f$ lies in a bounded convex subset of an RKHS defined by a positive definite kernel function $k(,)$, $\mathbf{x}_i$s and $y_i$s are the samples and their regression targets, respectively, and $\lambda_o\geq0$ is the regularization parameter. 

As a special case of ridge regression, correlation filters generate their training set $\{\mathbf{x}_i|i=0,\ldots,l-1\}$ by cyclically shifting a base sample, $\mathbf{x}\in \mathbb{R}^l$, such that $\mathbf{x}_i=\mathbf{P}^i_l\mathbf{x}$,
where $\mathbf{P}_l$ is the permutation matrix of $l\times l$~\cite{henriques15},
and the $y_i$s are often Gaussian labels.

By means of the Representer Theorem~\cite{scholkopf2002}, the optimal solution $f^*$ to Problem~(\ref{eq:tikhonov}) can be expressed as $f^*(\mathbf{x})=\sum_{i=0}^{l-1}\alpha_i k(\mathbf{x}_i,\mathbf{x})$.
Then, $||f||^2_k=\bm{\alpha}^{\top}\mathbf{K}\bm{\alpha}$, where $\bm{\alpha}=(\alpha_0,\alpha_1,\ldots,\alpha_{l-1})^{\top}$, and $\mathbf{K}$ is the positive semi-definite kernel matrix with $\kappa_{ij}=k(\mathbf{x}_i,\mathbf{x}_j)$ as its elements, and Problem~(\ref{eq:tikhonov}) becomes
\begin{equation}
\label{eq:krls}
	 \min_{\bm{\alpha}\in \mathbb{R}^l}\dfrac{1}{2}||\mathbf{y}-\mathbf{K}\bm{\alpha}||_2^2+\dfrac{\lambda_o}{2}\bm{\alpha}^{\top}\mathbf{K}\bm{\alpha}
\end{equation}
for $\bm{\alpha}$, where $\mathbf{y}=(y_0,y_1,\ldots,y_{l-1})^{\top}$.

It has been shown that using multiple kernels instead of a single one can improve the discriminability~\cite{lanc04b,varma2007}. Given the base kernels, $k_m$, where $m=1,2,\ldots,M$, a usual approach is to consider $k(\mathbf{x}_i,\mathbf{x}_j)$ to be a convex combination of base kernels, \ie, $k(\mathbf{x}_i,\mathbf{x}_j)=\mathbf{d}^{\top}\mathbf{k}(\mathbf{x}_i,\mathbf{x}_j)$, where $\mathbf{k}(\mathbf{x}_i,\mathbf{x}_j) = (k_1(\mathbf{x}_i,\mathbf{x}_j),k_2(\mathbf{x}_i,\mathbf{x}_j),\ldots,k_M(\mathbf{x}_i,\mathbf{x}_j))^{\top}$, $\mathbf{d}=(d_1,d_2,\ldots,d_M)^{\top}$, $\sum_{m=1}^M d_m=1$, and $d_m\geq0$. Hence we have $\mathbf{K}=\sum_{m=1}^M d_m\mathbf{K}_m$,
where $\mathbf{K}_m$ is the $m^{\text{th}}$ base kernel matrix with $\kappa_{ij}^m=k_m(\mathbf{x}_i,\mathbf{x}_j)$ as its elements. Substituting $\mathbf{K}$
for that in~(\ref{eq:krls}), we obtain the constrained optimization problem as follows.
\begin{equation}
\label{eq:mkrls}
\begin{array}{rl}
\displaystyle\min_{\bm{\alpha},\mathbf{d}} & F(\bm{\alpha},\mathbf{d}),\\
\textrm{s.t.} & \sum_{m=1}^{M}d_m=1,\\
& d_m\geq0,\;\;m=1,\ldots,M,
\end{array}
\end{equation}
where
\begin{equation}
\label{eq:mkrls-obj}
  F(\bm{\alpha},\mathbf{d})=\displaystyle\frac{1}{2}\left\|\mathbf{y}-\sum_{m=1}^M d_m\mathbf{K}_m\bm{\alpha}\right\|^2_2+
	 \dfrac{\lambda_o}{2}\bm{\alpha}^{\top}\sum_{m=1}^M d_m\mathbf{K}_m \bm{\alpha}.
\end{equation}
The optimal solution to Problem~(\ref{eq:mkrls}) can be expressed as
\begin{equation}
\label{eq:mkrepresenter}
	 f^*(\mathbf{x})=\sum_{i=0}^{l-1}\alpha_i\mathbf{d}^{\top}\mathbf{k}(\mathbf{x}_i,\mathbf{x}).
\end{equation}

Given $\mathbf{d}$ in Problem~(\ref{eq:mkrls}), we get an unconstrained quadratic programming problem \wrt $\bm{\alpha}$. And given $\bm{\alpha}$, Problem~(\ref{eq:mkrls}) is the constrained quadratic programming \wrt $\mathbf{d}$. Let $\{\mathbf{K}_m\}$ be positive semi-definite. Then, it is clear that given $\mathbf{d}$, $F(\bm{\alpha},\mathbf{d})$ is convex \wrt $\bm{\alpha}$, and given $\bm{\alpha}$, $F(\bm{\alpha},\mathbf{d})$ is convex \wrt $\mathbf{d}$.

To solve for $\bm{\alpha}$, let $\nabla_{\bm{\alpha}}F(\bm{\alpha},\mathbf{d})=0$; it is achieved that
\begin{equation}
\label{eq:mkrls_alpha}
\bm{\alpha}=\left(\sum_{m=1}^M d_m\mathbf{K}_m+\lambda_o\mathbf{I}\right)^{-1}\mathbf{y},
\end{equation}
where $\mathbf{I}$ is an $l\times l$ identity matrix. And $\mathbf{d}$ can be determined with the quadprog function in Matlab's optimization toolbox. Initially, $\forall m,\; d_m=1/M$. Then, because $F(\bm{\alpha},\mathbf{d})\geq0$, alternately evaluating Eq.~(\ref{eq:mkrls_alpha}) with fixed $\mathbf{d}$ and invoking the quadprog function with fixed $\bm{\alpha}$ for $\mathbf{d}$ will achieve a local optimal solution $(\bm{\alpha}^*, \mathbf{d}^*)$.

%
\subsubsection{Fast Evaluation in Training}
\label{sec:fasttraining}
As stated in Sec.~\ref{sec:mkcf}, the training samples are cyclically shifting in correlation filters. Therefore, the optimization processes of $\bm{\alpha}$ and $\mathbf{d}$ can be speeded up by means of the fast Fourier transform (FFT) pair, $\mathcal{F}$ and $\mathcal{F}^{-1}$.

At first, the evaluation of first rows $\mathbf{k}_{m}$s of kernel matrices $\mathbf{K}_m$s can be accelerated with FFT because the samples are circulant~\cite{henriques2012,henriques15}. Because $\mathbf{K}_m$s are circulant~\cite{henriques2012}, the inverses and the sum of circulant matrices are circulant~\cite{gray06}. Then the evaluation of Eq.~(\ref{eq:mkrls_alpha}) can be accelerated as
\begin{equation}
\label{eq:fft_alpha}
	\bm{\alpha}= \mathcal{F}^{-1}\left(\dfrac{\mathcal{F}(\mathbf{y})}{\mathcal{F}\left(\sum_{m=1}^M d_m\mathbf{k}_m\right)+\lambda_o}\right).
\end{equation}

According to Eq.~(\ref{eq:mkrls-obj}), given $\bm{\alpha}$, the optimization function $F(\mathbf{d};\bm{\alpha})$ \wrt $\mathbf{d}$ can be expressed as
\begin{equation}
\label{eq:mkrls-obj-d}
  F(\mathbf{d};\bm{\alpha})=\frac{1}{2}\mathbf{d}^{\top}\mathbf{A}_{\mathbf{d}}\mathbf{d}+\frac{1}{2}\mathbf{d}^{\top}\mathbf{B}_{\mathbf{d}}+\frac{1}{2}\mathbf{y}^{\top}\mathbf{y},
\end{equation}
where
\begin{equation}
\label{eq:mkrls-obj-d-A}
\mathbf{A}_{\mathbf{d}}=\left(
\begin{array}{ccc}
  \bm{\alpha}^{\top}\mathbf{K}_1^{\top}\mathbf{K}_1\bm{\alpha} & \cdots & \bm{\alpha}^{\top}\mathbf{K}_1^{\top}\mathbf{K}_M\bm{\alpha} \\
  \vdots & \ddots & \vdots \\
  \bm{\alpha}^{\top}\mathbf{K}_M^{\top}\mathbf{K}_1\bm{\alpha} & \cdots & \bm{\alpha}^{\top}\mathbf{K}_M^{\top}\mathbf{K}_M\bm{\alpha}
\end{array}
\right),
\end{equation}
and
\begin{equation}
\label{eq:mkrls-obj-d-B}
\mathbf{B}_{\mathbf{d}}=\left(\mathbf{b}_{\mathbf{d}}^{\top}\mathbf{K}_1\bm{\alpha},\ldots,\mathbf{b}_{\mathbf{d}}^{\top}\mathbf{K}_M\bm{\alpha}\right)^{\top},
\end{equation}
$\mathbf{b}_{\mathbf{d}}=\lambda_o\bm{\alpha}-2\mathbf{y}$. The evaluation of $\mathbf{A}_{\mathbf{d}}$ and $\mathbf{B}_{\mathbf{d}}$ can be accelerated by evaluating $\mathbf{K}_m\bm{\alpha}$ with $\mathcal{F}^{-1}(\mathcal{F}^{*}(\mathbf{k}_{m})\odot\mathcal{F}(\bm\alpha))$, where $m=1,\ldots,M$.

\subsubsection{Fast Detection}
\label{sec:detection}
According to Eq.~(\ref{eq:mkrepresenter}), the MKCF evaluates the responses of all test samples $\mathbf{z}_n=\mathbf{P}_l^n\mathbf{z}$, $n=0,1,\ldots,l-1$, in the current frame $p+1$ as
\begin{equation}
\label{eq:response}
y^n(\mathbf{z})=
\sum_{m=1}^M d_m\sum_{i=0}^{l-1}\alpha_i k_m(\mathbf{z}_n,\mathbf{x}^p_{m,i}),
\end{equation}
where $\mathbf{z}$ is the base test sample, $\mathbf{x}^p_{m,i}=\mathbf{P}^i_l\mathbf{x}^p_m$, $\mathbf{x}^p_m$ is the weighted average of the $m^{\mathrm{th}}$ feature of historical locations till frame $p$. Formally,
\begin{equation}
\label{eq:patchupdate}
\mathbf{x}^p_m=(1-\eta_m)\mathbf{x}^{p-1}_m+\eta_m R(D(\iota(p),s_p^*),\zeta,m),
\end{equation}
where $\eta_m\in[0,1]$ is the learning rate of kernel $m$ for the appearance of training samples, $\iota(p)$ and $s_p^*$ are the optimal location and scale of target object in frame $p$, respectively, $\zeta$ is the pre-defined scale for the image sequence, $D(\iota(p),s_p^*)$ is the image patch determined by $\iota(p)$ and $s_p^*$ in frame $p$, $R(D,\zeta,m)$ denotes $D$ re-sampled by $\zeta$ for kernel $m$, and $\mathbf{x}^0_m$ is the feature in the initial frame.

Because $k_m(,)$'s are permutation-matrix-invariant, the response map, $\mathbf{y}(\mathbf{z})$, of all virtual samples generated by $\mathbf{z}$ can be evaluated as
\begin{equation}
\label{eq:responsemap}
\mathbf{y}(\mathbf{z})\equiv(y^0(\mathbf{z}),\ldots,y^{l-1}(\mathbf{z}))^{\top}
=\sum_{m=1}^M d_m\mathbf{C}(\mathbf{k}^p_m)\alpha,
\end{equation}
where $\mathbf{k}^p_m=(k^p_{m,0},\ldots,k^p_{m,l-1})$, $k^p_{m,i}=k_m(\mathbf{z},\mathbf{P}^i_l\mathbf{x}^p_m)$, and $\mathbf{C}(\mathbf{k}_m^p)$ is the circulant matrix with $\mathbf{k}^p_m$ as its first row. Therefore, the response map can be accelerated as follows.
\begin{equation}
\label{eq:responsemapfft}
\mathbf{y}(\mathbf{z})=
\sum_{m=1}^M d_m\mathcal{F}^{-1}\left(\mathcal{F}^*(\mathbf{k}^p_m)\odot \mathcal{F}(\bm{\alpha})\right).
\end{equation}
The element of $\mathbf{y}({\mathbf{z}})$ which takes the maximal value is accepted as the optimal location of object in frame $p+1$. And the target's optimal scale is determined with fDSST~\cite{dane17a}.

\subsection{Shortcoming of Multi-kernel Correlation Filter}
\label{sec:mkcf-defect}
In order to achieve the robust performance of location, MKCF is updated with the weighted average of historical samples. To improve the location performance further, we would like to train a common MKCF (\ie, common $\bm{\alpha}$ and $\mathbf{d}$) for the historical samples, just like what was done in~\cite{dane14a}. Then, the optimization function should be as follows.
$$
\begin{array}{l}
F_e(\bm{\alpha},\mathbf{d})= \\
\displaystyle\sum_{j=1}^{p}\beta^j\left(\displaystyle\frac{1}{2}\left\|\mathbf{y}-\sum_{m=1}^M d_m\mathbf{K}_{m}^j\bm{\alpha}\right\|^2_2+
	 \dfrac{\lambda_o}{2}\bm{\alpha}^{\top}\sum_{m=1}^M d_m\mathbf{K}_{m}^j\bm{\alpha}\right)\\
=\displaystyle\frac{1}{2}\sum_{m=1}^{M}\sum_{j=1}^{p}\beta^j\left(\mathbf{y}^{\top}\mathbf{y}-2d_m\mathbf{y}^{\top}\mathbf{K}_{m}^j{\bm{\alpha}}+{\lambda_o}d_m\bm{\alpha}^{\top}
\mathbf{K}_{m}^j\bm{\alpha}\right)\\
+\displaystyle\frac{1}{2}\sum_{j=1}^{p}\beta^j{\bm{\alpha}}^{\top}\sum_{m=1}^{M}d_m\mathbf{K}_{m}^j\sum_{m=1}^{M}d_m\mathbf{K}_{m}^j\bm{\alpha},
\end{array}
$$
where $\beta^j$ is the weight of optimization function of the sample in frame $j$, $\mathbf{K}_{m}^j$ is the circulant kernel matrix with $\mathbf{k}_{m}^j$ as its first row, $\mathbf{k}^j_m=(k^j_{m,0},\ldots,k^j_{m,l-1})$, $k^j_{m,i}=k_m(\mathbf{z},\mathbf{P}_l^i\mathbf{x}^j_m)$, $j=1,\ldots,p$. $\mathbf{x}^j_m$ is evaluated by using Eq.~(\ref{eq:patchupdate}) where $j$ is used instead of $p$.

Commonly, different kernels (\i.e., features) should be equipped with different weights $\beta^j$, as their robustness is different throughout an image sequence. For example, the colors of the target object may vary more frequently than its HOG in an image sequence. Nevertheless, it is impossible for different kernels to set different $\beta^j$ in $F_e(\bm{\alpha},\mathbf{d})$, because different kernels are multiplied by each other and can not be separated into different items. Therefore, it is expectable that the location performance will be affected negatively if $F_e(\bm{\alpha},\mathbf{d})$, instead of $F(\bm{\alpha},\mathbf{d})$, is used in Problem~(\ref{eq:mkrls}), because different kernels have to share the same weight $\beta^j$.

\subsection{Extension of Multi-kernel Correlation Filter with Upper Bound}
\label{sec:emkcf}
Let $\mathbf{y}_c=\bf{y}/M$. We have
\[
\begin{split}
F(\bm{\alpha},\mathbf{d})&=\displaystyle\frac{1}{2}\left\|\mathbf{y}-\sum_{m=1}^M d_m\mathbf{K}_m\bm{\alpha}\right\|^2_2+
\dfrac{\lambda_o}{2}\bm{\alpha}^{\top}\sum_{m=1}^M d_m\mathbf{K}_m \bm{\alpha}\\
&\leq\displaystyle\frac{\mu}{2}\sum_{m=1}^M\left(\left\|\mathbf{y}_c-d_m\mathbf{K}_m\bm{\alpha}\right\|^2_2+
\lambda d_m\bm{\alpha}^{\top}\mathbf{K}_m \bm{\alpha}\right)\\
&\equiv U_{F(\bm{\alpha},\mathbf{d})},
\end{split}
\]
where $\mu=2M+1$, $\lambda=\frac{\lambda_o}{\mu}$, and the upper bound is reached when $d_{m_1}\mathbf{K}_{m_1}\bm{\alpha}=d_{m_2}\mathbf{K}_{m_2}\bm{\alpha}$, $m_1=1,\ldots,M$ and $m_2=1,\ldots,M$. The proof can be found in the supplementary material. We then treat $U_{F(\bm{\alpha},\mathbf{d})}$, the upper bound of $F(\bm{\alpha},\mathbf{d})$, as the optimization function of MKCF and introduce the historical samples into it. Consequently, the final optimization objective for training a common multi-kernel correlation filter for the whole historical samples can be expressed as follows.
$$
F_p(\bm{\alpha}_p,\mathbf{d}_p)\equiv\displaystyle\frac{1}{2}\sum_{j=1}^p\sum_{m=1}^M\beta^j_{m}u^{j,m}_{F(\bm{\alpha},\mathbf{d})},
$$
where
$$
u^{j,m}_{F(\bm{\alpha},\mathbf{d})}=\left\|\mathbf{y}_c-d_{m,p}\mathbf{K}^j_{m}\bm{\alpha}_p
\right\|^2_2+\lambda d_{m,p}\bm{\alpha}_p^{\top}\mathbf{K}^j_{m}\bm{\alpha}_p,
$$
$\beta^1_{m}=(1-\gamma_m)^{p-1}$, $\beta^j_{m}=\gamma_m(1-\gamma_m)^{p-j}$, $j=2,\ldots,p$, $p$ is the number of historical frames, $\gamma_m\in(0,1)$ is the learning rate of kernel $m$ for the common MKCF, $\mathbf{K}^j_m$ is the Gram matrix of the $m^{\text{th}}$ kernel for the samples in frame $j$, $\bm{\alpha}_p=(\alpha_{0,p},\alpha_{1,p},\ldots,\alpha_{l-1,p})^{\top}$ and $\mathbf{d}_p=(d_{1,p},d_{2,p},\ldots,d_{M,p})^{\top}$ are dual vector and weight vector of all kernels when frame $p$ is processed, respectively, and $\sum_{m=1}^{M}d_{m,p}=1$. And the new optimization problem for the MKCF with whole samples is
\begin{equation}
\label{eq:emkrls}
\begin{array}{rl}
\displaystyle\min_{\bm{\alpha}_p,\mathbf{d}_p} & F_p(\bm{\alpha}_p,\mathbf{d}_p),\\
\textrm{s.t.} & \sum_{m=1}^{M}d_{m,p}=1,\\
& d_{m,p}\geq0,\;\;m=1,\ldots,M.
\end{array}
\end{equation}
This is a constrained optimization problem. And similar to Problem~(\ref{eq:mkrls}), given $\mathbf{d}_p$, $F_p(\bm{\alpha}_p,\mathbf{d}_p)$ is convex and unconstrained \wrt $\bm{\alpha}_p$, and given $\bm{\alpha}_p$, $F_p(\bm{\alpha}_p,\mathbf{d}_p)$ is convex and constrained \wrt $\mathbf{d}_p$.

Because $F_p(\bm{\alpha}_p,\mathbf{d}_p)$ is unconstrained \wrt $\bm{\alpha}_p$, to solve for $\bm{\alpha}_p$, let $\nabla_{\bm{\alpha}_p}F_p(\bm{\alpha}_p,\mathbf{d}_p)=0$; we achieve that
\begin{equation}
\label{eq:alpha-emkcf}
\begin{split}
\bm{\alpha}_p= & \left(\displaystyle\sum_{j=1}^p\sum_{m=1}^M\beta^j_{m}\left((d_{m,p}\mathbf{K}_{m}^j)^2+\lambda d_{m,p}\mathbf{K}_{m}^j\right)\right)^{-1}\cdot\\
& \displaystyle\sum_{j=1}^p\sum_{m=1}^M\beta_{m}^j d_{m,p}\mathbf{K}_{m}^j\mathbf{y}_c,
\end{split}
\end{equation}
which can be evaluated efficiently with FFT as follows.
\begin{equation*}
\begin{split}
\mathcal{A}_p&\equiv\mathcal{F}(\bm{\alpha}_p)\\
&=\dfrac{\displaystyle\sum_{j=1}^{p}\sum_{m=1}^{M}\beta_{m}^j\mathcal{F}(d_{m,p}\mathbf{k}_{m}^j)\odot
  \mathcal{F}(\mathbf{y}_c)}{\displaystyle\sum_{j=1}^{p}\sum_{m=1}^{M}\beta_{m}^j\mathcal{F}(d_{m,p}\mathbf{k}_{m}^j)
  \odot(\mathcal{F}(d_{m,p}\mathbf{k}_{m}^j)+\lambda)}.\\
\end{split}
\end{equation*}
Set
\begin{equation}
\label{eq:alpha-evaluation}
\mathcal{A}_p=\frac{\mathcal{A}_p^N}{\mathcal{A}_p^D}=\frac{\sum_{m=1}^{M}\mathcal{A}_{m,p}^N}{\sum_{m=1}^{M}\mathcal{A}_{m,p}^D},
\end{equation}
where
\[
\begin{split}
&\mathcal{A}_{m,p}^N=(1-\gamma_m)\mathcal{A}_{m,p-1}^N+\gamma_m\mathcal{F}(d_{m,p}\mathbf{k}_{m}^p)\odot
\mathcal{F}(\mathbf{y}_c),\\
&\begin{split}
\mathcal{A}_{m,p}^D= & (1-\gamma_m)\mathcal{A}_{m,p-1}^D+\\
                     & \gamma_m\mathcal{F}(d_{m,p}\mathbf{k}_{m}^p)\odot(\mathcal{F}(d_{m,p}\mathbf{k}_{m}^p)+\lambda),
\end{split}
\end{split}
\]
if $p>1$. In the initial frame, $p=1$. Then
\[
\begin{split}
&\mathcal{A}_{m,1}^N=\mathcal{F}(d_{m,1}\mathbf{k}_{m}^1)\odot\mathcal{F}(\mathbf{y}_c),\\
&\mathcal{A}_{m,1}^D=\mathcal{F}(d_{m,1}\mathbf{k}_{m}^1)\odot(\mathcal{F}(d_{m,1}\mathbf{k}_{m}^1)+\lambda).
\end{split}
\]
Therefore, $\mathcal{A}_p$ can be evaluated efficiently frame by frame.

Solving for $\mathbf{d}_p$ in Problem~(\ref{eq:emkrls}) will have to deal with a constrained optimization problem. This means that it is difficult to obtain an iteration scheme for the optimal $\mathbf{d}_p^*$ which is as efficient as the one for $\bm{\alpha}_p^*$. Now let us investigate the constraints in Problem~(\ref{eq:emkrls}). It is clear that there are three purposes for adding these constraints in Problem~(\ref{eq:emkrls}). (1) $d_{m,p}\geq0$, $m=1,\ldots,M$, are necessary to ensure $\sum_{m=1}^{M}d_{m,p}$ is convex combination. (2) $\sum_{m=1}^{M}d_{m,p}=1$ is necessary to ensure the optimal $\mathbf{d}_p^*$ is unique and its value is finite. (3) Both $d_{m,p}\geq0$ and $\sum_{m=1}^{M}d_{m,p}=1$ are necessary to ensure there exists at least an $m$ such that $d_{m,p}>0$. Therefore, if we are able to design an algorithm to optimize the unconstrained problem
\begin{equation}
\label{eq:emkrls-2}
\displaystyle\min_{\bm{\alpha}_p,\mathbf{d}_p}F_p(\bm{\alpha}_p,\mathbf{d}_p)
\end{equation}
\wrt $\mathbf{d}_p$, such that the above three requirements are satisfied implicitly,
then the explicit constraints in Problem~(\ref{eq:emkrls}) can be canceled. In the rest of this section, we will first derive an efficient algorithm to optimize Problem~(\ref{eq:emkrls-2}) \wrt $\mathbf{d}_p$, and then prove that the optimal $\mathbf{d}_p^*$ indeed implicitly satisfies the above requirements for the optimal solution if $d_{m,1}>0$, $m=1,\ldots,M$.

To solve for $\mathbf{d}_p$ in Problem~(\ref{eq:emkrls-2}), let $\nabla_{\mathbf{d}_p}F_p(\bm{\alpha}_p,\mathbf{d}_p)=0$. Then, it is achieved that
$$
d_{m,p}=\frac{\sum_{j=1}^{p}\beta_{m}^j(\mathbf{K}_{m}^j\bm{\alpha}_p)^\top(2\mathbf{y}_c-\lambda\bm{\alpha}_p)}
{2\sum_{j=1}^{p}\beta_{m}^j(\mathbf{K}_{m}^j\bm{\alpha}_p)^\top(\mathbf{K}_{m}^j\bm{\alpha}_p)},
$$
where $m=1,\ldots,M$.
Set
\begin{equation}
\label{eq:d-evaluation}
d_{m,p}=\frac{d_{m,p}^N}{d_{m,p}^D},
\end{equation}
where
\[
\begin{split}
&d_{m,p}^N=(1-\gamma_m)d_{m,p-1}^N+\gamma_m(\mathbf{K}_{m}^p\bm{\alpha}_p)^\top(2\mathbf{y}_c-\lambda\bm{\alpha}_p),\\
&d_{m,p}^D=(1-\gamma_m)d_{m,p-1}^D+2\gamma_m(\mathbf{K}_{m}^p\bm{\alpha}_p)^\top(\mathbf{K}_{m}^p\bm{\alpha}_p),
\end{split}
\]
if $p>1$. And if $p=1$, then
\[
\begin{split}
&d_{m,1}^N=(\mathbf{K}_{m}^1\bm{\alpha}_1)^\top(2\mathbf{y}_c-\lambda\bm{\alpha}_1),\\
&d_{m,1}^D=2(\mathbf{K}_{m}^1\bm{\alpha}_1)^\top(\mathbf{K}_{m}^1\bm{\alpha}_1).
\end{split}
\]
It is clear that $\mathbf{K}_{m}^p\bm{\alpha}_p$ can be accelerated with
$$
\mathcal{F}^{-1}(\mathcal{F}^{*}(\mathbf{k}_{m}^p)\odot\mathcal{F}(\bm{\alpha}_p))
=\mathcal{F}^{-1}(\mathcal{F}^{*}(\mathbf{k}_{m}^p)\odot\mathcal{A}_p).
$$
Therefore, $d_{m,p}$ can be evaluated efficiently, and optimal solution $\mathbf{d}_p^*$ can be obtained efficiently frame by frame.

\begin{theorem}
\label{th:positive-d}
Suppose that $\mathbf{K}_{m}^j$ is circulant Gram matrix, $\lambda>0$, all components of $\mathbf{y}_c$ is positive, and also suppose $d_{m,p}^t>0$, $m=1,\ldots,M$, $j=1,\ldots,p$, $t=1,2,\ldots$, where $d_{m,p}^t$ is the $t^{\emph{th}}$ iteration on frame $p$ when solving Problem~(\ref{eq:emkrls-2}) with alternative evaluation of $\bm{\alpha}_p$ and $\mathbf{d}_p$. Then,
\begin{itemize}
  \item[(1)] $d_{m,p}^{t+1}>0$,
  \item[(2)] 
            $c_l\cdot\lambda/2+c_l\cdot b^{\min}<d_{m,p}^{t+1}<c_u\cdot\lambda/2+c_u\cdot b^{\max}$,
  where $c_l$ and $c_u$ are two constants determined by $\mathbf{y}_c$, discrete Fourier transform matrix, $\beta^j_m$, and the eigenvalues of $\mathbf{K}_{m}^j$, $b^{\min}$ and $b^{\max}$ are two constants related to $d_{m,p}^t$, $\beta^j_m$, and the eigenvalues of $\mathbf{K}_{m}^j$.
\end{itemize}
\end{theorem}
The proof can be found in the supplementary material.

It can be seen from Theorem~\ref{th:positive-d} that the range of $d_{m,p}^{t+1}$ is totally determined by two lines \wrt $\lambda$ when $d_{m,p}^1$ is fixed. The smaller $\lambda$, the smaller $d_{m,p}^{t+1}$, therefore, the smaller the components of final optimal solution $\mathbf{d}_p^*$. That is, the components of $\mathbf{d}_p^*$ are always finite and controlled by $\lambda$. It is obvious that $\mathbf{d}_p^*$ satisfies the three requirements for the optimal solution of Problem~(\ref{eq:emkrls-2}) \wrt $\mathbf{d}_p$, given the initial $d_{m,p}^1>0$, $m=1,\ldots,M$.

More refined analysis on the relationship of $\lambda$ and optimal $\mathbf{d}_p^*$ is complex, because the bounds of $\mathbf{d}_p^*$ heavily depend on the eigenvalues of all kernel matrices which are constructed with practical samples and an additional scale parameter in the kernel. Therefore, we will experimentally show the further numerical relation between $\lambda$ and $\mathbf{d}_p^*$ in Sec.~\ref{sec:d-lambda}.

Based on the above analysis, it is concluded that the optimization objective of the extension of MKCF is Problem~(\ref{eq:emkrls-2}), and its optimization process is as follows. Initially, $d_{m,1}=1/M$, $m=1,\ldots,M$. Then alternately evaluate Eq.~(\ref{eq:alpha-evaluation}) with fixed $\mathbf{d}_p$ and Eq.~(\ref{eq:d-evaluation}) with fixed $\bm{\alpha}_p$. Because $F_p(\bm{\alpha}_p,\mathbf{d}_p)\geq0$ is convex \wrt $\bm{\alpha}_p$ and $\mathbf{d}_p$, respectively, such iterations will converge to a local optimal solution $(\bm{\alpha}_p^*, \mathbf{d}_p^*)$. In our experiments, a satisfactory convergency $(\bm{\alpha}_p^*, \mathbf{d}_p^*)$ on frame $p$ can be achieved in three iterations of Eq.~(\ref{eq:alpha-evaluation}) and Eq.~(\ref{eq:d-evaluation}).

The fast determination of the optimal location and scale of target object in frame $p+1$ is the same as that of MKCF described in Sec.~\ref{sec:detection}, where $\bm{\alpha}=\bm{\alpha}_p^*$ and $\mathbf{d}=\mathbf{d}_p^*$.

\section{Implementation Details}
\label{sec:details}
In our experiments, the color and HOG are used as features in MKCFup. Considering the tradeoff between the discriminability and computational cost, we employ a kernel for each of color and HOG, \ie, $M=2$. As in~\cite{dane14a,henriques15,dane15b,tangm15}, the multiple channels of the color and HOG are concatenated into a single vector, respectively. The response map $\mathbf{y}$ is identical to that in KCF~\cite{henriques2012}.

The color scheme proposed by~\cite{dane14a} is adopted as our color feature, except that we reduce the dimensionality of color to four with principal component analysis (PCA). Normal nine gradient orientations and $4\times4$ cell size are utilized in HOGs. The dimensionality of our HOGs is also reduced to four with PCA to speed up MKCFup. Gaussian kernel is used for both features with $\sigma_{\mathrm{color}}=0.515$ and $\sigma_{\mathrm{HOG}}=0.6$ for color sequences and $\sigma_{\mathrm{color}}=0.3$ and $\sigma_{\mathrm{HOG}}=0.4$ for gray sequences. 
Employing Gaussian kernel to construct kernel matrices ensures that all $\mathbf{K}_m$s are positive definite~\cite{micchelli86}. The learning rates $\gamma_{\mathrm{color}}=0.0174$ and $\gamma_{\mathrm{HOG}}=0.0173$ for color sequences, and $\gamma_{\mathrm{color}}=0.0175$ and $\gamma_{\mathrm{HOG}}=0.018$ for gray sequences. The learning rates of sample appearance $\eta_{\mathrm{color}}=\gamma_{\mathrm{color}}$ and $\eta_{\mathrm{HOG}}=\gamma_{\mathrm{HOG}}$ for both color and gray sequences.

In order to reduce high-frequency noise in the frequency domain stemming from the large discontinuity between opposite edges of a cyclic-extended image patch, the feature patches are banded with Hann window. Because there is only one true sample in each frame, it is well known that too large a search region in KCF will reduce the location performance~\cite{henriques2012,dane14a}. Therefore, the search region is set 2.5 times larger than the bounding box of target object, which is the same as that in KCF and $\text{CN}_2$~\cite{dane14a}.

\vspace{-2mm}
\section{Experimental Results}
\label{sec:exper}
The MKCFup was implemented in MATLAB. The experiments were performed on a PC with Intel Core i7 3.40GHz CPU and 8GB RAM.

It is well-known that all samples of MOSSE, KCF, MKCF, and MKCFup are circulant. Therefore, their search region can not be set too large~\cite{dane15a}. Too large a search region will include too much background, significantly reducing the discriminability of filters for target object against background. Consequently, the search regions of above CF trackers have to be set experientially around 2.5 times larger than the object bounding boxes~\cite{henriques15,tangm15}, much smaller than those of CFLB, SRDCF, and ECO\_HC~\cite{galoo15,dane15a,dane17b}. It is obvious that it will be impossible for any tracker to catch the target object once the target moves out of its search region in the next frame. Therefore, CFLB, SRDCF, and ECO\_HC are better for locating the target object of large move than KCF, MKCF, and MKCFup.

An even worse situation for KCF, MKCF, and MKCFup is that, according to the experimental experiences on correlation filter based trackers~\cite{bolme10,henriques2012,dane14b,tangm15}, even if the target is in the search region in next frame, its location may still be unreliable when the target moving near to the boundaries of the search region. Specifically, it is often difficult for the CF trackers, such as MOSSE, $\text{CN}_2$~\cite{dane14a}, KCF, MKCF, and MKCFup which use only one base sample, to obtain a reliable location by using response maps if the ratio of the center distance of target object over the bounding box in two frames is larger than 0.6 when the background clutter is present. Consequently, it is suitable for the above CF trackers to track the target object with quite small move between two frames. In this paper, the move of target object is defined as \emph{small}, if the offset ratio
\begin{equation}
\label{eq:largemovestd}
\tau\equiv\frac{\|\mathbf{c}(\mathbf{x}_t)-\mathbf{c}(\mathbf{x}_{t+\delta})\|_2}{\sqrt{w(\mathbf{x}_t)\cdot h(\mathbf{x}_t)}}<0.6,
\vspace{-2mm}
\end{equation}
where $\mathbf{c}()$, $w()$, and $h()$ are the center, width, and height of sample, respectively. $\delta=1$ if there is no occlusion for the target object, otherwise $\delta$ is the amount of frames from starting to ending occlusion. A sequence is accepted to contain the target object of large move if there exists two adjacent frames or the occlusion of target object such that $\tau>0.6$. It is noted that the above definition of offset ratio for small move is quite rough, because it neglects the possible big difference between width and height.

According to the above discussion, two visual tracking benchmarks, OTB2013~\cite{wu13} and NfS~\cite{galoo17b} were utilized to compare different trackers in this paper, because most of sequences of OTB2013 and most of high frequency part of NfS only contain small move of the target object.

In our experiments, the trackers are evaluated in one-pass evaluation (OPE) using both precision and success plots~\cite{wu13}, calculated as percentages of frames with center errors lower than a threshold and the intersection-over-union (IoU) overlaps exceeding a threshold, respectively. Trackers are ranked using the precision score with center error lower than 20 pixels and area-under-the-curve (AUC), respectively, in precision and success plots.

In this paper, to simplify the experiments, we only compare those state-of-the-art trackers which merely employ the hand-crafted features color or HOG.

\subsection{Relationship of optimal weight $\mathbf{d}_p^*$ and regularization parameter $\lambda$}
\label{sec:d-lambda}

\begin{figure}[t]
	\centering
	\includegraphics[width=3.2in]{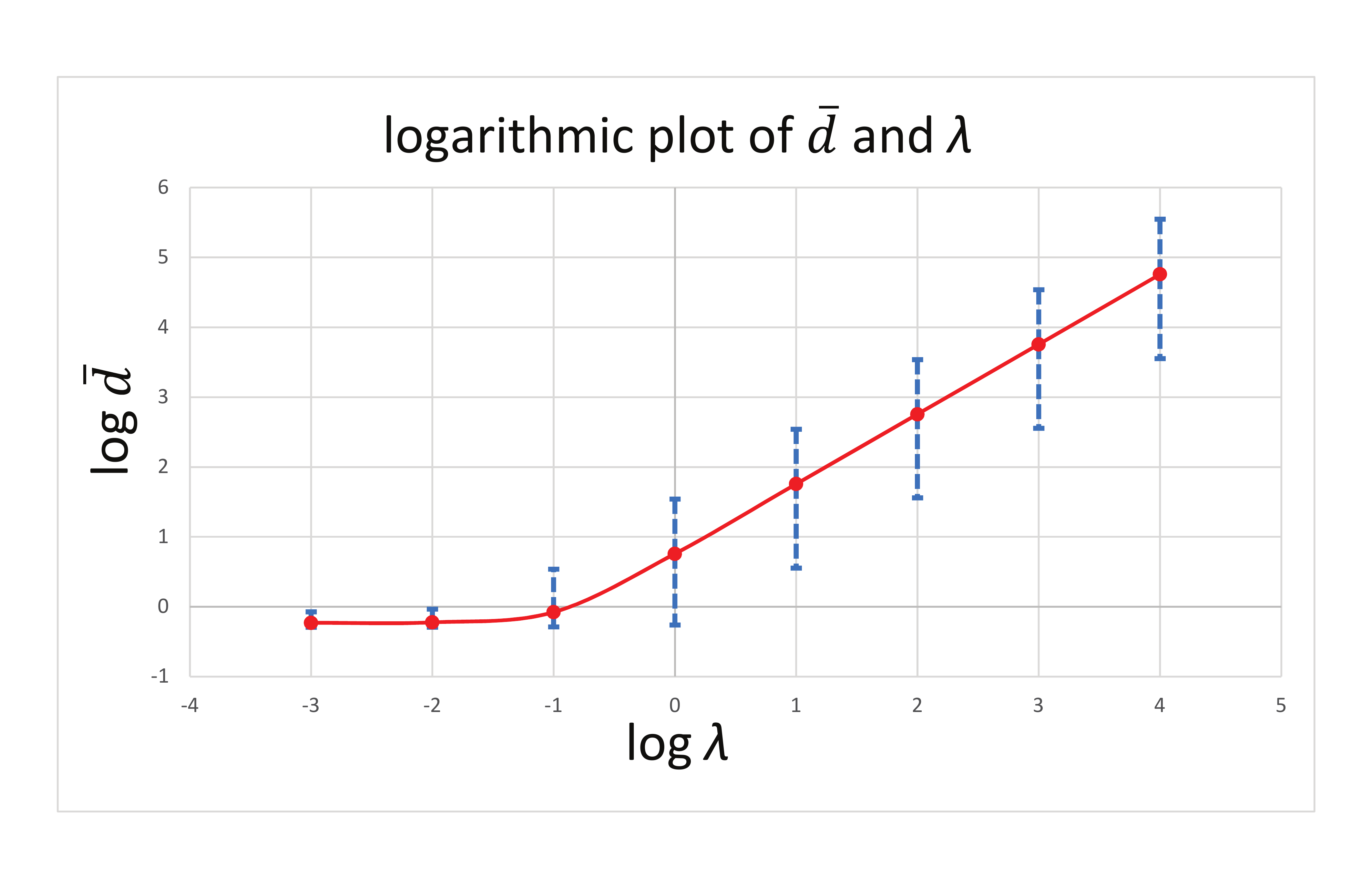}
	\caption{The numerical relationship between regularization parameter $\lambda$ and $\overline{d}$, the average of $d_{m,p}^*$ over $m$ on OTB2013~\cite{wu13}. Besides $\overline{d}$, two deviations away from $\overline{d}$ are also presented. The logarithmic function is employed to make the relation more clear. See Sec.~\ref{sec:d-lambda} for details.}
	\label{fig:lambda-d}
\end{figure}

Fig.~\ref{fig:lambda-d} shows the numerical relation of $\lambda$ and $\mathbf{d}_p^*$ obtained on OTB2013 when initially $\mathbf{d}_{p}^1=(0.5,0.5)$. In the experiment, $\lambda\in\{10^{-3},10^{-2},10^{-1},1,10,10^2,10^3,10^4\}$. According to Theorem~\ref{th:positive-d}, we set $d_{m,p}^*=d_{m,p}^*(\lambda)$, because $d_{m,p}^*$ is a function of $\lambda$, and
\[
\begin{split}
&\overline{d}(\lambda)=\frac{1}{MPS}\sum_m\sum_p\sum_i d_{m,p,i}^*(\lambda),\\
&(\delta_{\max}(\lambda), \delta_{\min}(\lambda))=(\max_{m,p,i}d_{m,p,i}^*(\lambda), \min_{m,p,i}d_{m,p,i}^*(\lambda)),
\end{split}
\]
where $P$ and $S$ are the number of selected frames in each image sequence and the total number of selected sequences, respectively, $p$ and $i$ represent the number of selected frame and the number of selected sequence, respectively, and $d_{m,p,i}^*$ is the optimal weight of the $m^{\text{th}}$ kernel at frame $p$ of sequence $i$. In our experiment, specifically, $P=10$ and $S=20$. That is, for each $\lambda$, ten frames are randomly sampled from each of 20 randomly selected sequences out of OTB2013, and $d_{m,p}^*(\lambda)$s on these frames are used to calculate $\overline{d}(\lambda)$ and two deviations, $\delta_{\max}(\lambda)$ and $\delta_{\min}(\lambda)$. To demonstrate the relationship more clearly, $\lambda$ and its three functions are shown with logarithmic function.

It is interesting to notice that the relation of the averages of $\lambda$ and optimal $\mathbf{d}_{p}^*$ is almost linear when $\lambda<10^{-1}$ or $\lambda\geq1$. And $\delta_{\max}(\lambda)$ and $\delta_{\min}(\lambda)$ drop significantly when $\lambda<10^{-1}$. When $\lambda\leq0.05$, the deviations are really close to the average, and the relation of $\lambda$ and $\mathbf{d}_{p}^*$ itself is approximately linear. Surprisingly, $\frac{1}{M}\sum_{m=1}^M d_{m,p}^*\approx0.5$ for the frames of all sequences when $\lambda<10^{-1}$ in our experiment. That is, $\sum_{m=1}^M d_{m,p}^*\approx1$, because $M=2$. This means that the constraint of Problem~(\ref{eq:emkrls}) on the sum of all components of the optimal $\mathbf{d}_p$ is satisfied implicitly and approximately, while optimizing Problem~(\ref{eq:emkrls-2}) \wrt $\mathbf{d}_p$ with the iterations of Eq.~(\ref{eq:alpha-evaluation}) and Eq.~(\ref{eq:d-evaluation}).

\subsection{Comparison among MKCFs}
\label{sec:3mkcfcomp}
\begin{figure}
	\centering
	\includegraphics[width=0.235\textwidth]{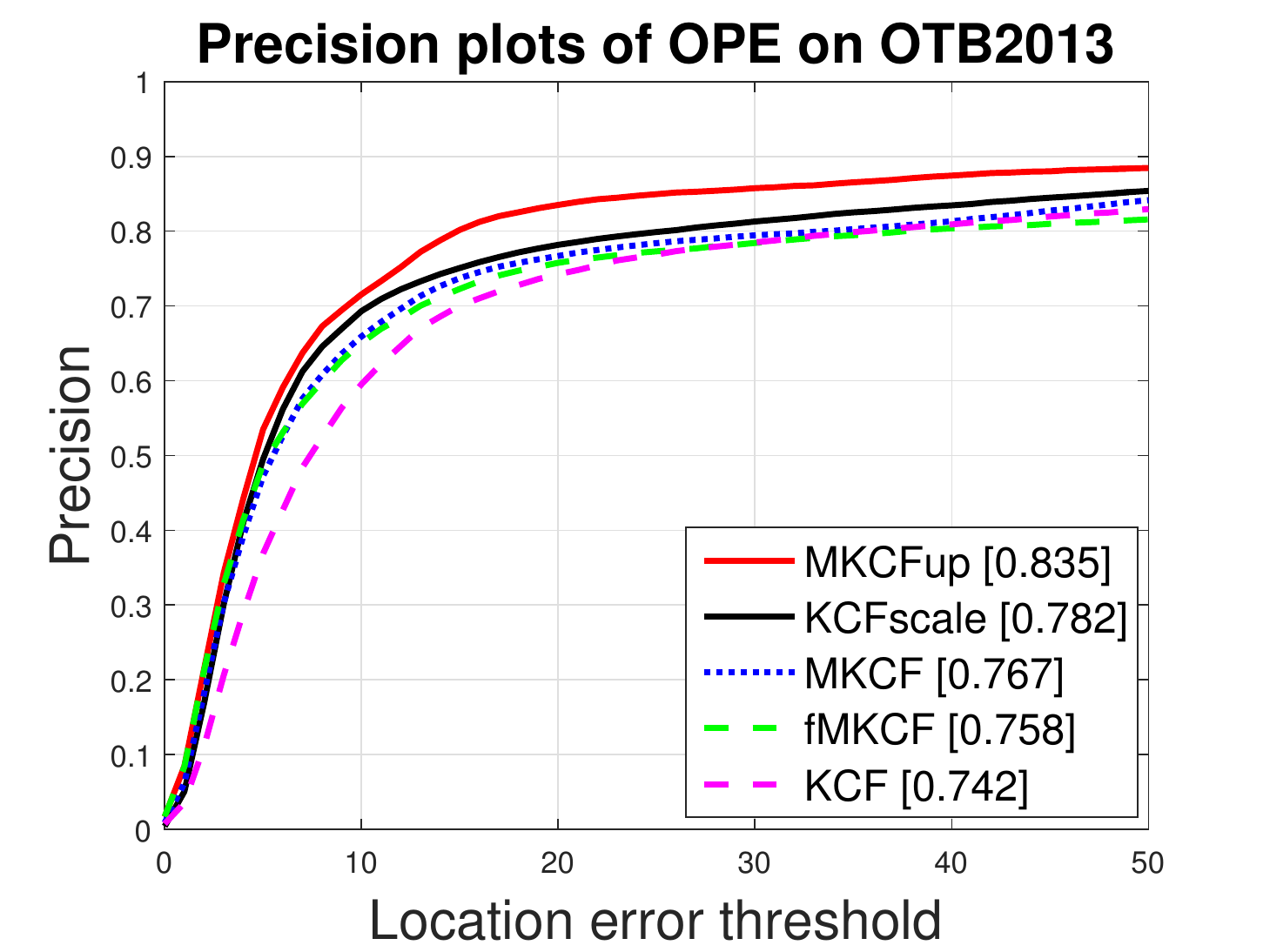}
	\includegraphics[width=0.235\textwidth]{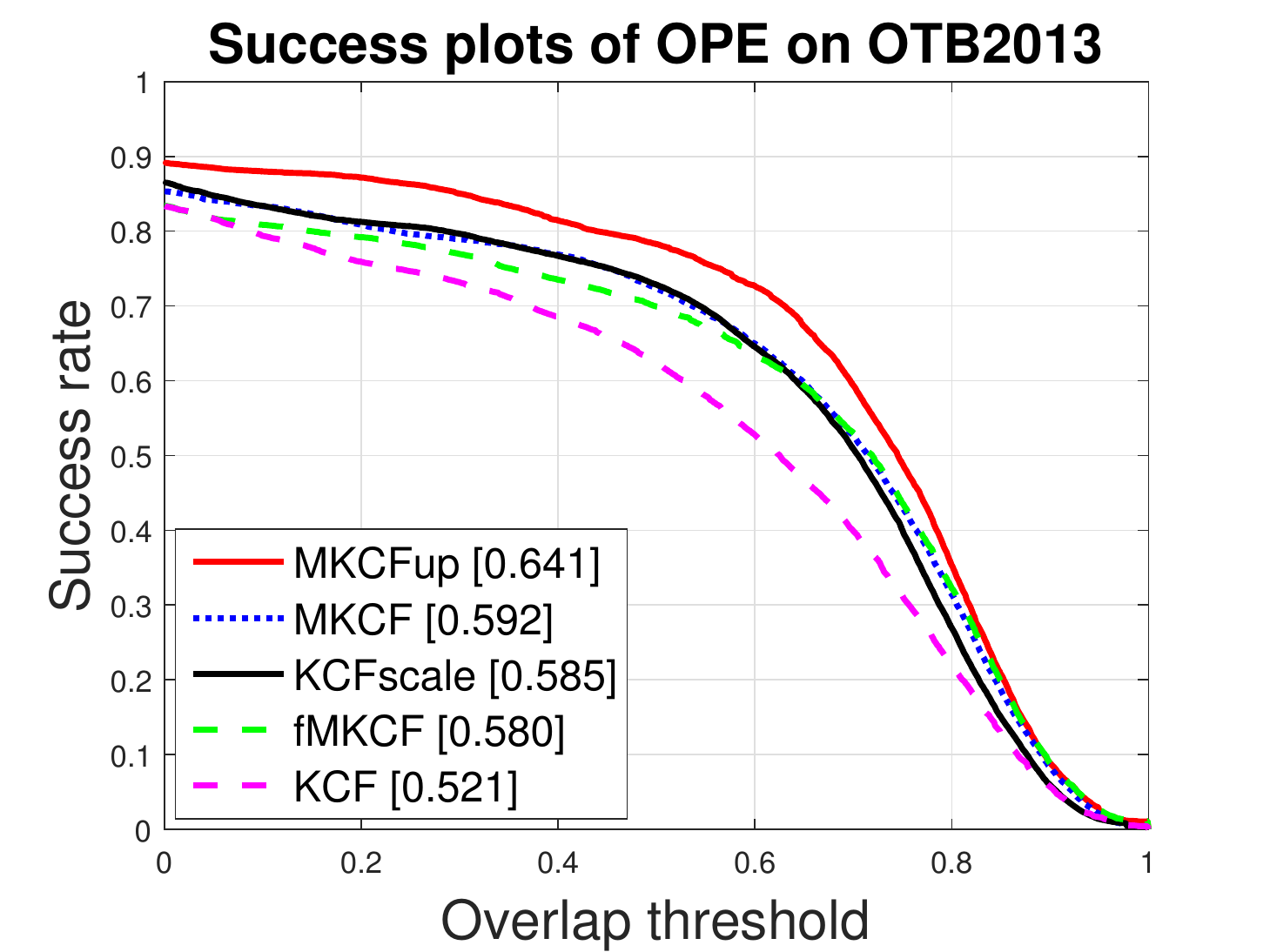}
	\caption{The precision and success plots of KCF~\cite{henriques15}, KCFscale, MKCF~\cite{tangm15}, fMKCF, and MKCFup on OTB2013~\cite{wu13}. See Sec.~\ref{sec:3mkcfcomp} for details. The average precision scores and AUCs of the trackers on the sequences are reported in the legends.}
	\label{fig:3mkcfcomp}
\end{figure}

In this section, we consider KCF as a special case of the original MKCF~\cite{tangm15} with $M=1$. To verify our improvement on KCF and MKCF is effective, we compare KCF, KCFscale, MKCF, fMKCF, and MKCFup on OTB2013, where KCFscale is the KCF with the scaling scheme of patch pyramid, and fMKCF is a variant of MKCF whose features and scaling scheme are the same as those adopted by MKCFup, and the optimization of $\mathbf{d}$ that is more efficient than the one in~\cite{tangm15}, as described in Sec.~\ref{sec:fasttraining}, is adopted. Fig.~\ref{fig:3mkcfcomp} reports the results. It is concluded from the figure that MKCFup outperforms KCF and KCFscale with large margins in both center precision and IoU, and that the novel objective function and training scheme of MKCFup improve the location performance with the average precision score of 83.5\% and the average AUC score of 64.1\%, significantly outperforming MKCF and fMKCF by 6.8\% and 4.9\% and 7.7\% and 6.1\%, respectively. It is noticed that the location performances of fMKCF are inferior to those of MKCF, although its fps is higher than MKCF's (50 vs. 30).

\subsection{Comparison to State-of-the-art Trackers with Handcrafted Features}
\label{sec:comphf}
\begin{figure*}
	\centering
	\includegraphics[width=0.24\textwidth]{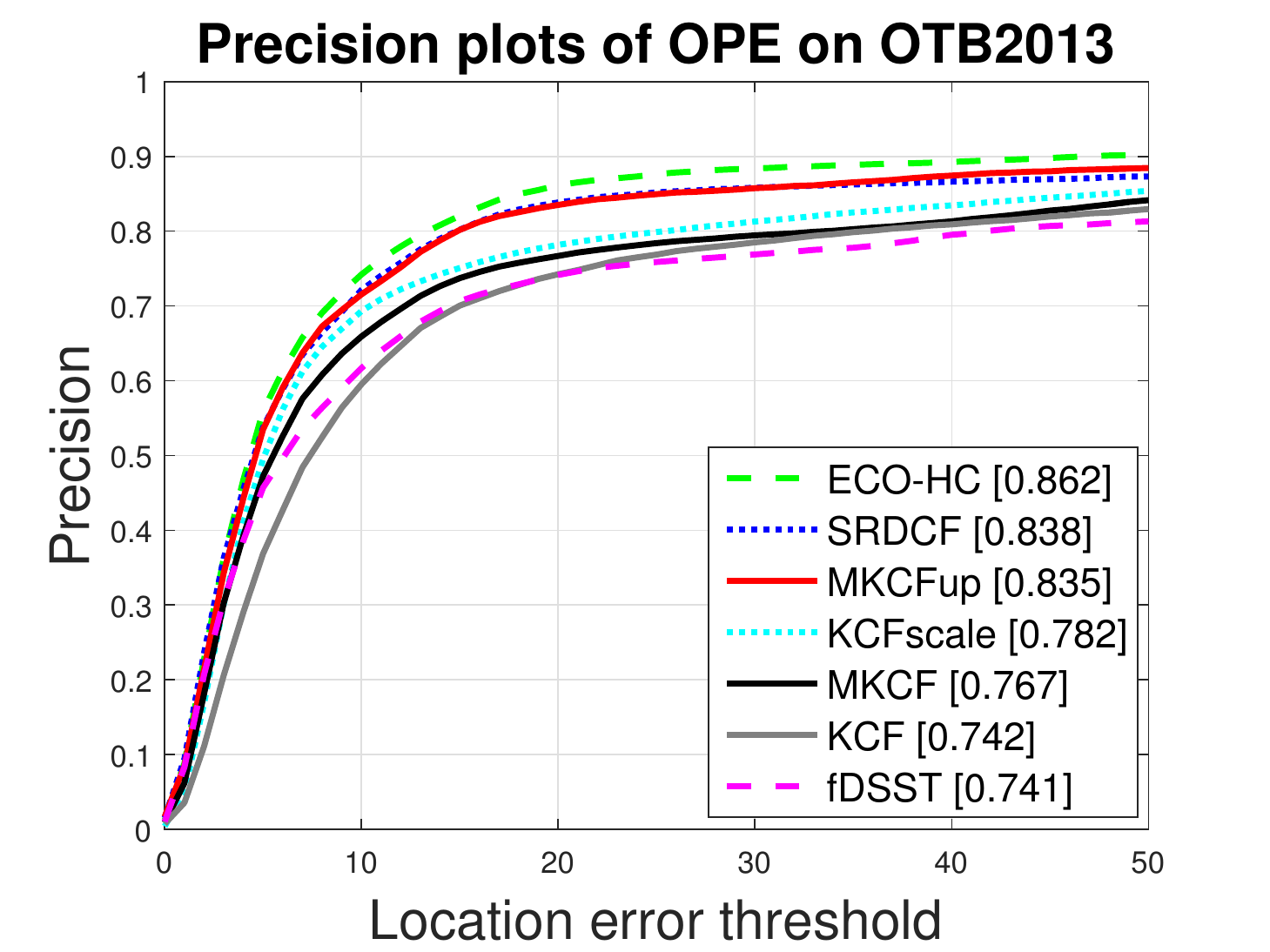}
	\includegraphics[width=0.24\textwidth]{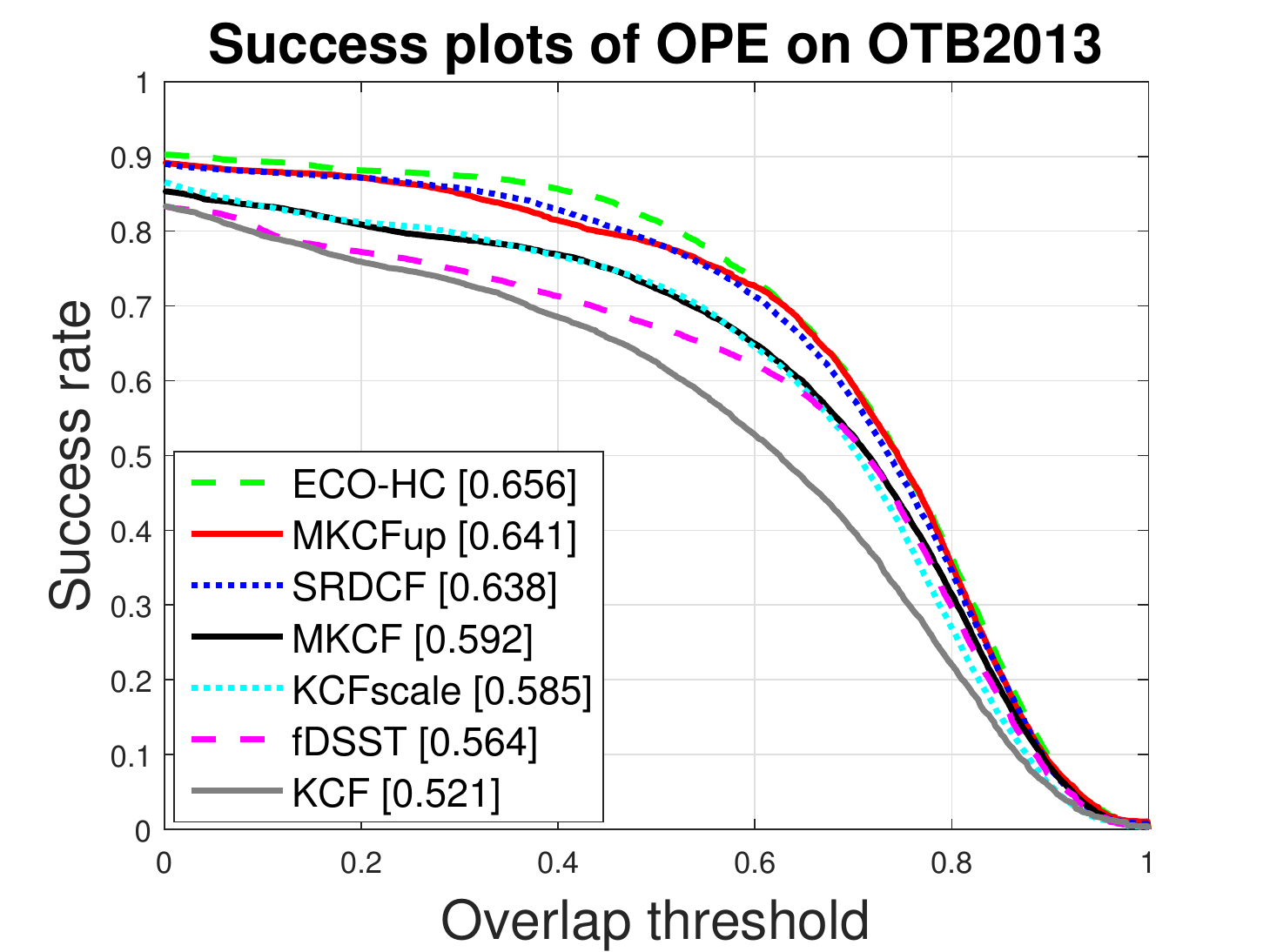}
	\includegraphics[width=0.24\textwidth]{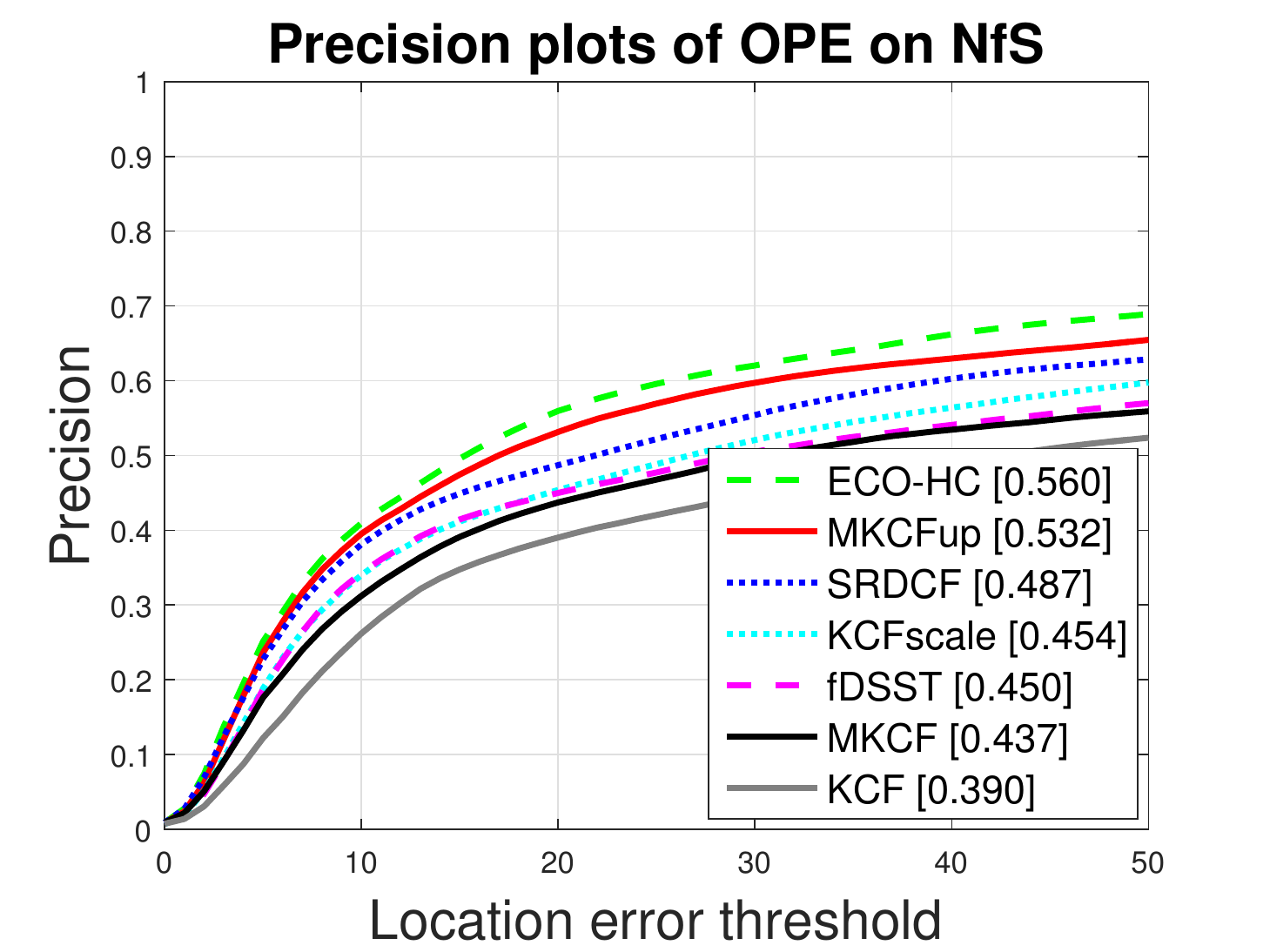}
	\includegraphics[width=0.24\textwidth]{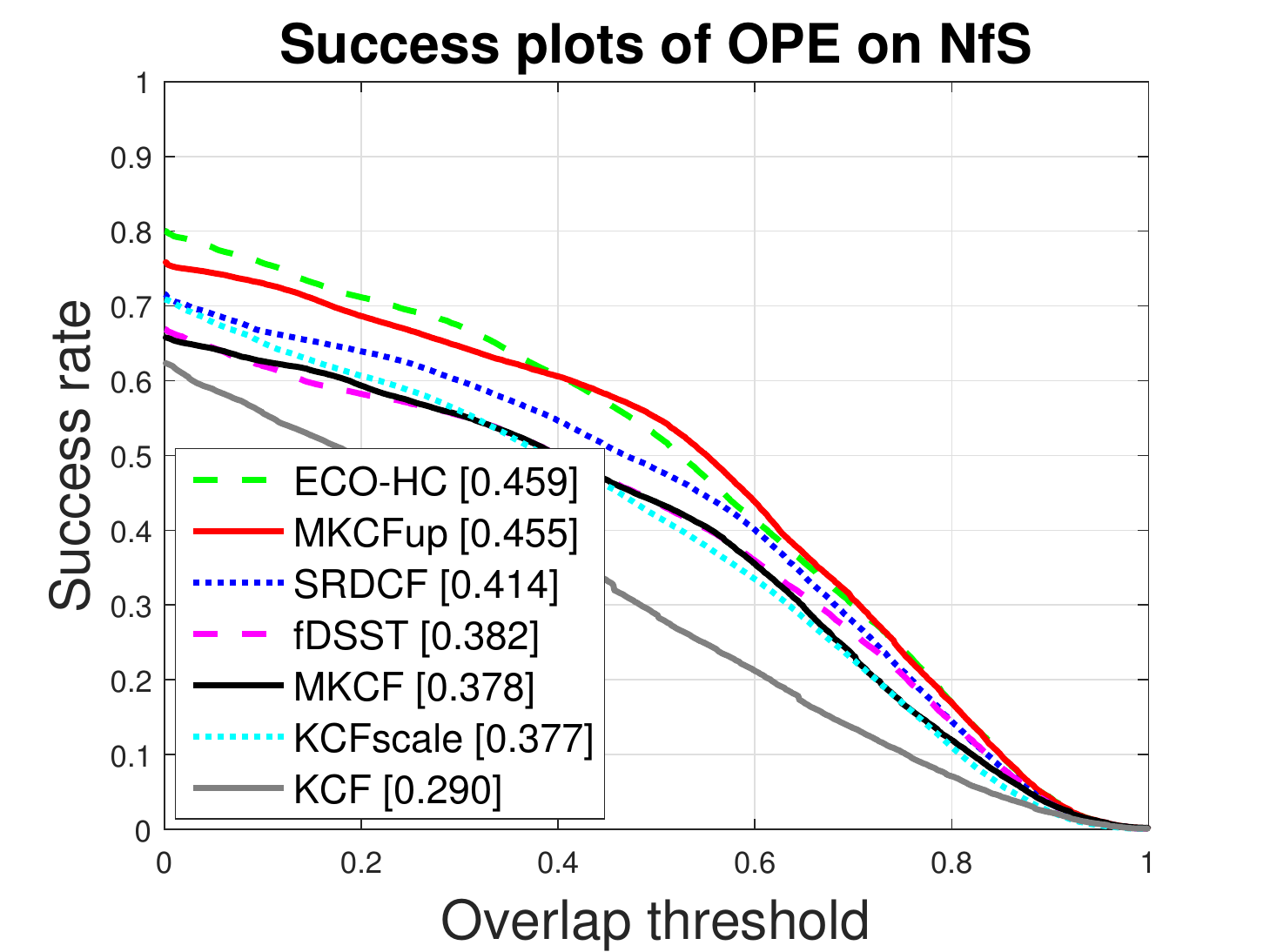}
	\caption{The precision and success plots of MKCFup, KCF, KCFscale, MKCF, SRDCF, fDSST, and ECO\_HC on OTB2013~\cite{wu13} and NfS~\cite{galoo17b}. The average precision scores and AUCs of the trackers on the sequences are reported in the legends.}
	\label{fig:comphf}
\end{figure*}

We compare our MKCFup to other 6 trackers, KCF, KCFscale, MKCF, SRDCF, fDSST, and ECO\_HC on OTB2013 and NfS. Fig.~\ref{fig:comphf} shows the results. It can be seen that MKCFup outperforms all other trackers in both precision scores and AUCs, except for ECO\_HC, on two benchmarks. ECO\_HC is able to exploit larger search regions than MKCFup does to catch the target object of large move, whereas MKCFup is not. Therefore, ECO\_HC outperforms MKCFup on the whole benchmarks.

\subsection{Comparison on Sequences of Small Move}
\label{sec:largemovement}
\begin{figure}
	\centering
	\includegraphics[width=0.235\textwidth]{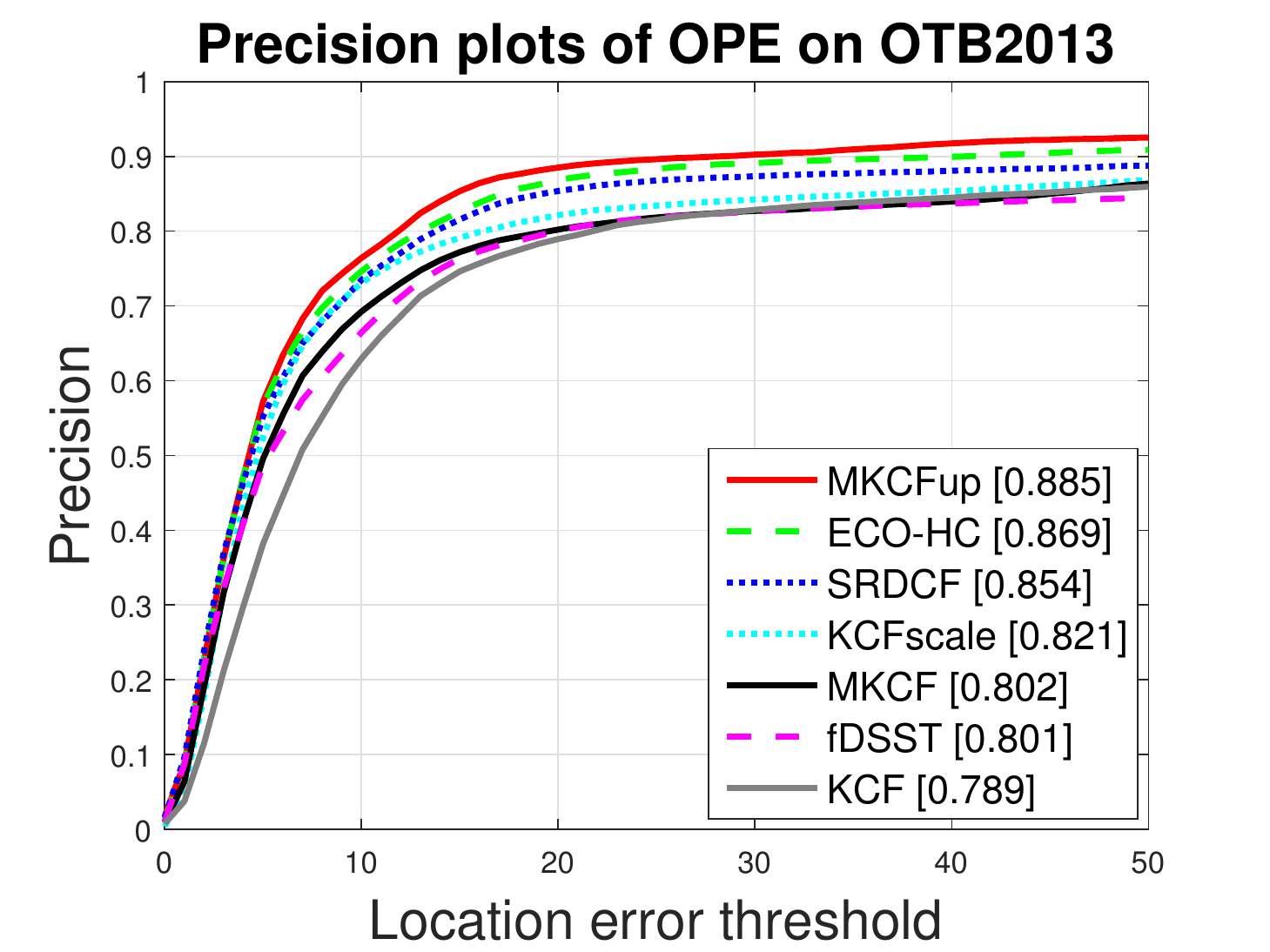}
	\includegraphics[width=0.235\textwidth]{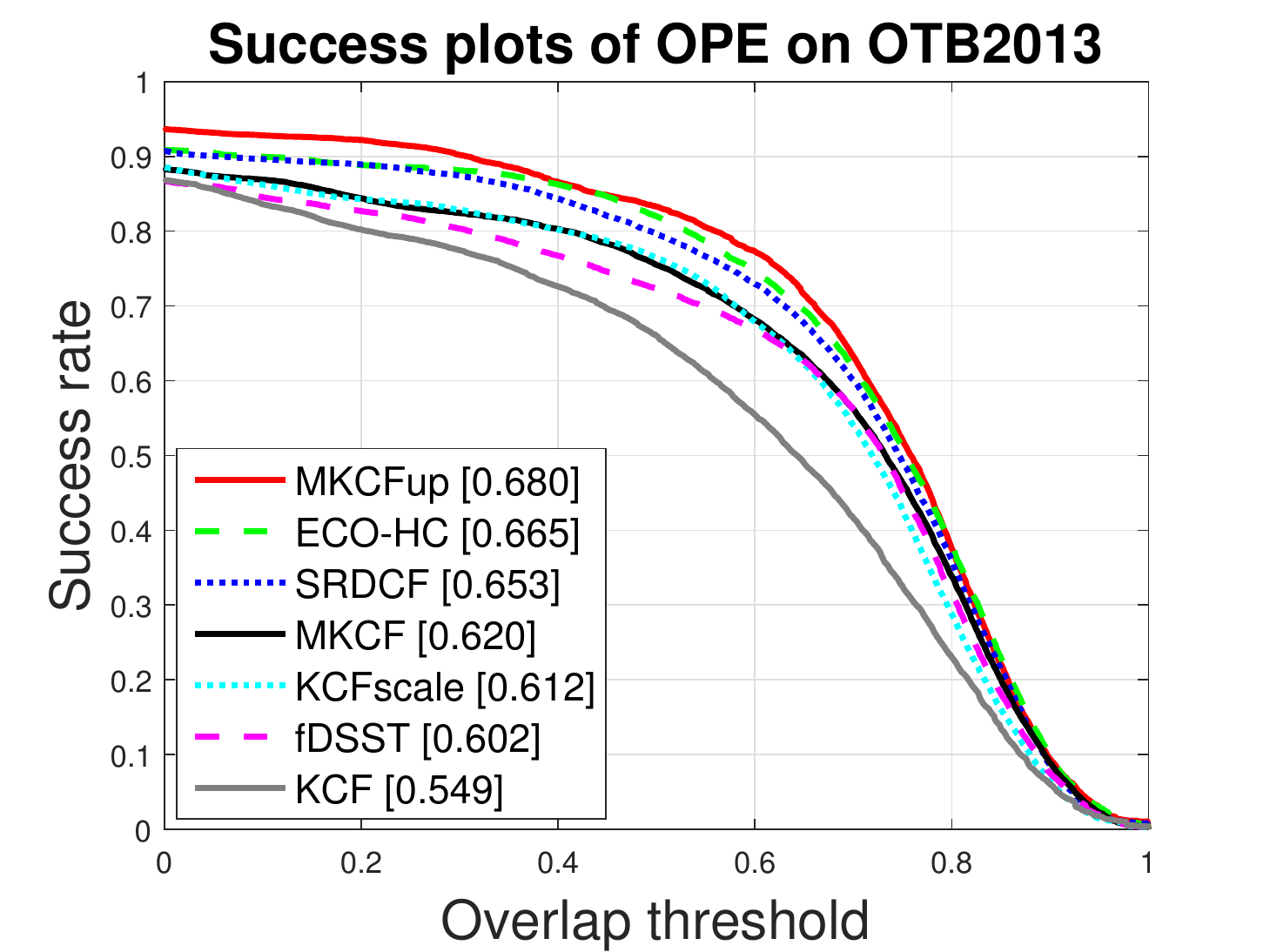}
	\caption{The precision and success plots of MKCFup, KCF, KCFscale, MKCF, SRDCF, fDSST, and ECO\_HC on small move sequences of OTB2013~\cite{wu13}. The average precision scores and AUCs of the trackers on the sequences are reported in the legends.}
	\label{fig:otb2013smallcomparison}
\end{figure}

\begin{figure}
	\centering
	\includegraphics[width=0.235\textwidth]{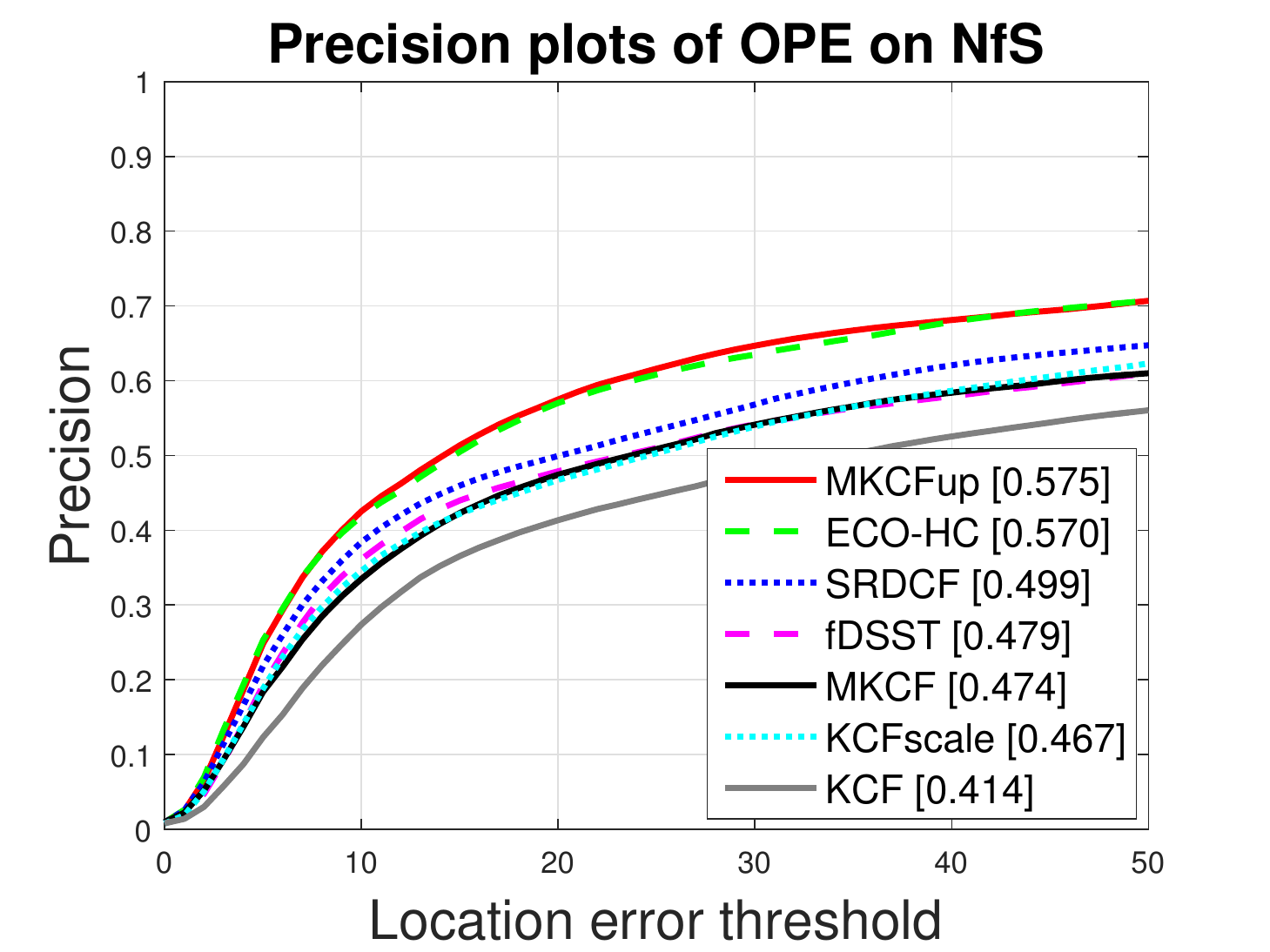}
	\includegraphics[width=0.235\textwidth]{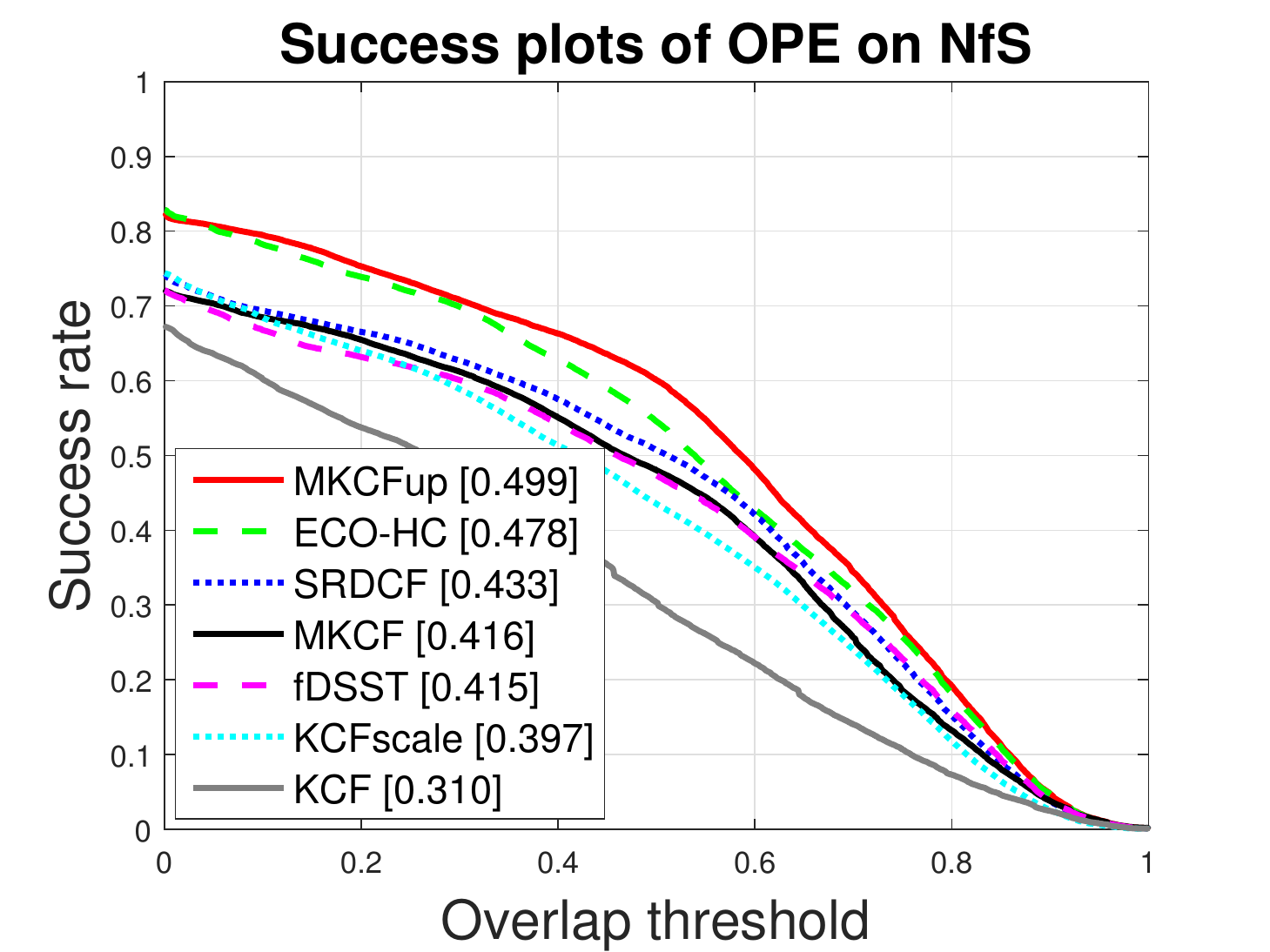}
	\caption{The precision and success plots of MKCFup, KCF, KCFscale, MKCF, SRDCF, fDSST, and ECO\_HC on small move sequences of NfS~\cite{galoo17b}. The average precision scores and AUCs of the trackers on the sequences are reported in the legends.}
	\label{fig:NfSsmallcomparison}
\end{figure}

By means of Eq.~(\ref{eq:largemovestd}), it is found that there exist six sequences which contain large move in OTB2013.\footnote{The 6 sequences which contain the target object with large move in OTB2013 are boy, matrix, tiger2, ironman, couple, jumping.} We then removed them from the benchmark and compared our MKCFup, KCF, KCFscale, MKCF, SRDCF, fDSST, and ECO\_HC on the rest sequences. Fig.~\ref{fig:otb2013smallcomparison} reports the results. It is seen that MKCFup outperforms SRDCF and ECO\_HC on the average precision score and AUC by $3.1\%$ and $2.7\%$ and $1.6\%$ and $1.5\%$, respectively, on the small move sequences of OTB2013.


To verify the advantage of MKCFup further, we removed the large move sequences\footnote{The 16 sequences which contain the target object with large move in NfS are airboard\_1, airtable\_3, bee, bowling3, football\_skill, parkour, pingpong8, basketball\_1, basketball\_3, basketball\_6, bowling2, dog\_2, pingpong2, motorcross, person\_scooter, soccer\_player\_3.} from NfS by means of Eq.~(\ref{eq:largemovestd}), and compared the above trackers on the rest 84 sequences. Note that an occluded target object is considered undergoing large move if its $\tau>0.6$ between two frames of starting and ending occlusion. Fig.~\ref{fig:NfSsmallcomparison} shows the results. It is seen that MKCFup outperforms SRDCF and ECO\_HC on the average precision score and AUC by $7.6\%$ and $6.6\%$ and $0.5\%$ and $2.1\%$, respectively, on small move sequences of NfS.



Table~\ref{tab:fps} lists the amount of frames the trackers can process per second.
\begin{table}\tiny{
	\caption{The amount of frames processed per second (fps) with different trackers.}
	\begin{center}
		\begin{tabular}{|c|c|c|c|c|c|c|c|}
			\hline
			Tracker & KCF & MKCF & fMKCF & fDSST & SRDCF & ECO-HC & MKCFup\\
			\hline\hline
			fps & 297 & 30 & 50 & 80 & 6 & 39 & 150\\
			\hline
		\end{tabular}
	\end{center}
\label{tab:fps}}
\vspace{-4mm}
\end{table}

According to the above experiments, it can be concluded that MKCFup outperforms state-of-the-art trackers, such as SRDCF and ECO\_HC, with much higher fps as long as the move of target object is small.

\section{Conclusions and Future Work}
\label{sec:conclusion}
A novel tracker, MKCFup, has been presented in this paper. By optimizing the upper bound of the objective function of original MKCF and introducing the historical samples into the upper bound, we derived the novel MKCFup. It has been demonstrated that the discriminability of MKCFup is more powerful than those of state-of-the-art trackers, such as SRDCF and ECO-HC, although its search region is much smaller than theirs. And the MKCFup'fps is much larger than state-of-the-art trackers'. In conclusion, MKCFup outperforms state-of-the-arts trackers with handcrafted features at high speed if the target object moves small.

{\small
\bibliographystyle{ieee}
\bibliography{bibjournal}
}

\end{document}


\title{\texttt{Supplementary Material}\\
\vskip 4mm
High-speed Tracking with Multi-kernel Correlation Filters}

\author{Ming Tang$^{1,2}$, Bin Yu$^{1,2}$, Fan Zhang$^3$, and Jinqiao Wang$^{1,2}$\\
$^1$University of Chinese Academy of Sciences, Beijing, China\\
$^2$National Lab of Pattern Recognition, Institute of Automation, CAS, Beijing 100190, China\\
$^3$School of Info. \& Comm. Eng., Beijing University of Posts and Telecommunications
}

\maketitle

\begin{abstract}
This supplementary material includes 1) the experimental results on OTB2015, 2) the proof of Theorem 1.
\vspace{-6mm}
\end{abstract}

\section{Experimental Results on OTB2015}
\begin{figure}[h]
	\centering
	\includegraphics[width=0.35\textwidth]{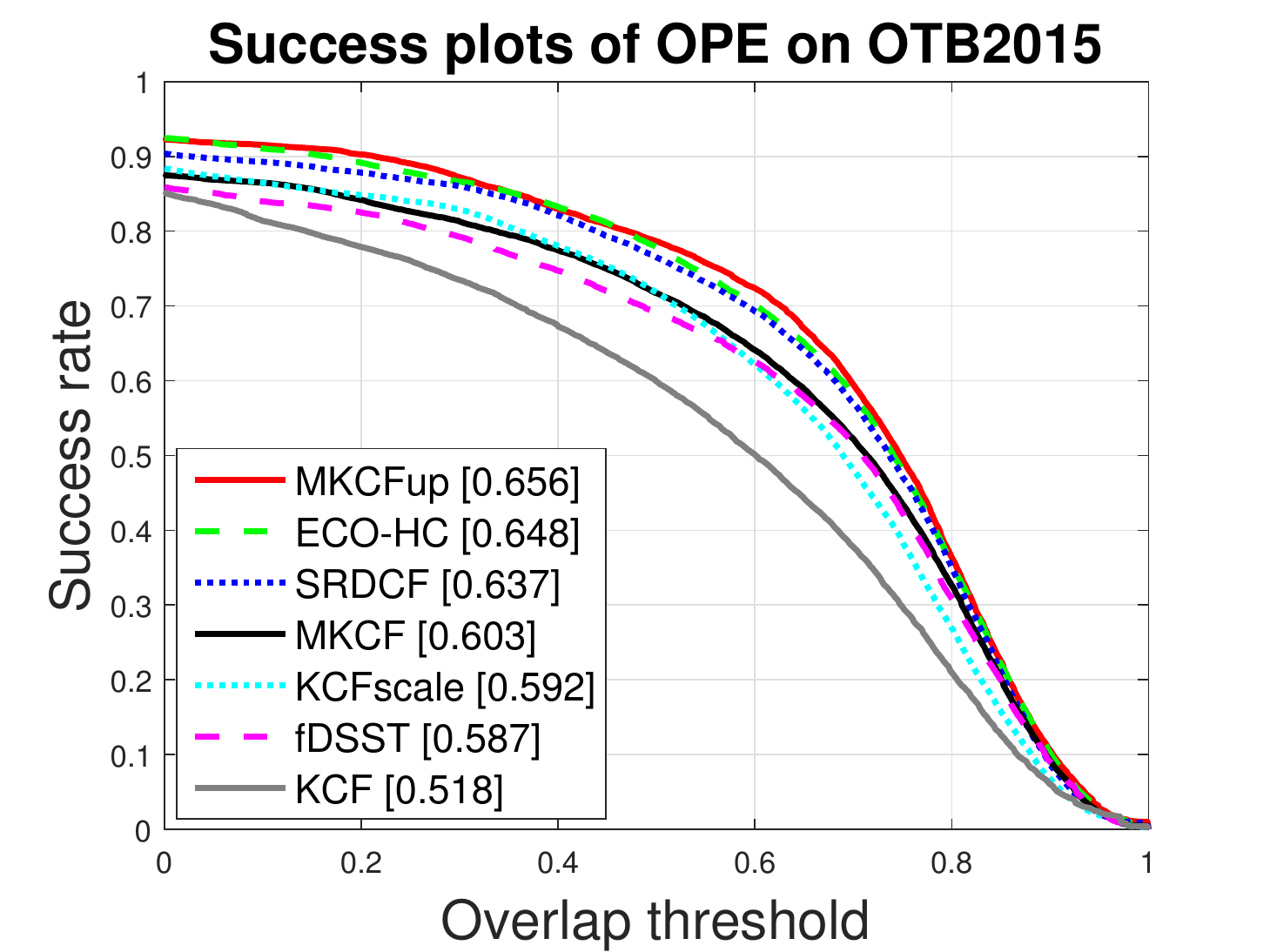}
	\caption{The success plot of MKCFup, KCF, KCFscale, MKCF, SRDCF, fDSST, and ECO\_HC on small move sequences of OTB2015. The AUCs of the trackers on the sequences are reported in the legends.}
	\label{fig:NfSsmallcomparison}
\vspace{-3mm}
\end{figure}

\section{Proof of Upper Bound of MKCF Objective Function~\cite{tangm15}}
\begin{lemma}
\label{lemma:1}
Suppose $\mathbf{a}_m$ is a vector, $m=1,\dots,M$. Then,
\[
\left\|\sum_{m=1}^{M}\mathbf{a}_m\right\|_2^2\leq(2M+1)\sum_{m=1}^{M}\|\mathbf{a}_m\|_2^2.
\]
\end{lemma}
The equality holds when $\mathbf{a}_m=\mathbf{a}_n$, where $m=1,\ldots,M$ and $n=1,\ldots,M$.

\noindent Proof of Lemma~\ref{lemma:1}:

It is true that
\[
\begin{split}
& \left\|\displaystyle\sum_{m=1}^{M}\mathbf{a}_m\right\|_2^2=\left(\displaystyle\sum_{m=1}^{M}
\mathbf{a}_m\right)^{\top}\left(\displaystyle\sum_{m=1}^{M}\mathbf{a}_m\right)=\\
&\displaystyle\sum_{m=1}^{M}\mathbf{a}_m^{\top}\displaystyle\sum_{m=1}^{M}\mathbf{a}_m=
\displaystyle\sum_{m=1}^{M}\mathbf{a}_m^{\top}\mathbf{a}_m+2\displaystyle\sum_{m=1}^{M}\sum_{n=1}^{M}
\mathbf{a}_m^{\top}\mathbf{a}_n.
\end{split}
\]

\noindent It is also true that
\[
\begin{split}
& \mathbf{a}_m^{\top}\mathbf{a}_m+\mathbf{a}_n^{\top}\mathbf{a}_n-2\mathbf{a}_m^{\top}\mathbf{a}_n=\\
& \mathbf{a}_m^{\top}(\mathbf{a}_m-\mathbf{a}_n)-\mathbf{a}_n^{\top}(\mathbf{a}_m-\mathbf{a}_n)=\\
& (\mathbf{a}_m-\mathbf{a}_n)^{\top}(\mathbf{a}_m-\mathbf{a}_n)=\|\mathbf{a}_m-\mathbf{a}_n\|_2^2\geq0.
\end{split}
\]
Therefore,
\[
\begin{split}
& \mathbf{a}_m^{\top}\mathbf{a}_m+\mathbf{a}_n^{\top}\mathbf{a}_n\geq2\mathbf{a}_m^{\top}\mathbf{a}_n,\\
& \sum_{m=1}^{M}\mathbf{a}_m^{\top}\mathbf{a}_m+M\mathbf{a}_n^{\top}\mathbf{a}_n\geq2
\sum_{m=1}^{M}\mathbf{a}_m^{\top}\mathbf{a}_n,\\
& M\sum_{m=1}^{M}\mathbf{a}_m^{\top}\mathbf{a}_m+M\sum_{n=1}^{M}\mathbf{a}_n^{\top}\mathbf{a}_n\geq2
\sum_{n=1}^{M}\sum_{m=1}^{M}\mathbf{a}_m^{\top}\mathbf{a}_n,\\
& 2M\displaystyle\sum_{m=1}^{M}\mathbf{a}_m^{\top}\mathbf{a}_m\geq2\sum_{m=1}^{M}\sum_{n=1}^{M}
\mathbf{a}_m^{\top}\mathbf{a}_n.
\end{split}
\]
Therefore,
\[
\begin{split}
   \left\|\displaystyle\sum_{m=1}^{M}\mathbf{a}_m\right\|_2^2 & =\displaystyle\sum_{m=1}^{M}\mathbf{a}_m^{\top}
   \mathbf{a}_m+2\displaystyle\sum_{m=1}^{M}\sum_{n=1}^{M}\mathbf{a}_m^{\top}\mathbf{a}_n\\
   & \leq\displaystyle\sum_{m=1}^{M}\mathbf{a}_m^{\top}\mathbf{a}_m+2M\displaystyle\sum_{m=1}^{M}
     \mathbf{a}_m^{\top}\mathbf{a}_m\\
   & =(2M+1)\displaystyle\sum_{m=1}^{M}\mathbf{a}_m^{\top}\mathbf{a}_m.
\end{split}
\]
That is,
\[
\left\|\displaystyle\sum_{m=1}^{M}\mathbf{a}_m\right\|_2^2\leq(2M+1)\displaystyle\sum_{m=1}^{M}
\|\mathbf{a}_m\|_2^2.
\]
It is clear that the equality holds when $\mathbf{a}_m=\mathbf{a}_n$, where $m=1,\ldots,M$ and $n=1,\ldots,M$.

\vspace{3mm}
\noindent Q.E.D.

\subsection{Proof of Upper Bound}
According to Lemma~\ref{lemma:1},
\[
\left\|\mathbf{y}-\sum_{m=1}^M d_m\mathbf{K}_m\bm{\alpha}\right\|^2_2\leq(2M+1)\sum_{m=1}^M
\left\|\mathbf{y}_c-d_m\mathbf{K}_m\bm{\alpha}\right\|^2_2.
\]
Therefore, $U_{F(\bm{\alpha},\mathbf{d})}$ is the upper bound of $F(\bm{\alpha},\mathbf{d})$, and the upper bound is reached when $d_m\mathbf{K}_m\bm{\alpha}=d_n\mathbf{K}_n\bm{\alpha}$, where $m=1,\ldots,M$ and $n=1,\ldots,M$.

\vspace{3mm}
\noindent Q.E.D.

\section{Proof of Theorem 1}
In the extension of MKCF with upper bound, to optimize the unconstrained problem
\begin{equation}
\label{eq:emkrls-2}
\displaystyle\min_{\bm{\alpha}_p,\mathbf{d}_p}F_p(\bm{\alpha}_p,\mathbf{d}_p),
\end{equation}
we achieve that
\begin{equation}
\label{eq:alpha-emkcf}
\begin{split}
\bm{\alpha}_p= & \left(\displaystyle\sum_{j=1}^p\sum_{m=1}^M\beta^j_{m}\left((d_{m,p}\mathbf{K}_{m}^j)^2+\lambda d_{m,p}\mathbf{K}_{m}^j\right)\right)^{-1}\cdot\\
& \displaystyle\sum_{j=1}^p\sum_{m=1}^M\beta_{m}^j d_{m,p}\mathbf{K}_{m}^j\mathbf{y}_c,
\end{split}
\end{equation}
and
\begin{equation}
\label{eq:d-evaluation}
d_{m,p}=\frac{d_{m,p}^N}{d_{m,p}^D},
\end{equation}
where
\[
\begin{split}
&d_{m,p}^N=(1-\gamma_m)d_{m,p-1}^N+\gamma_m(\mathbf{K}_{m}^p\bm{\alpha}_p)^\top(2\mathbf{y}_c-\lambda\bm{\alpha}_p),\\
&d_{m,p}^D=(1-\gamma_m)d_{m,p-1}^D+2\gamma_m(\mathbf{K}_{m}^p\bm{\alpha}_p)^\top(\mathbf{K}_{m}^p\bm{\alpha}_p),
\end{split}
\]
when $p>1$. If $p=1$, then
\[
\begin{split}
&d_{m,1}^N=(\mathbf{K}_{m}^1\bm{\alpha}_1)^\top(2\mathbf{y}_c-\lambda\bm{\alpha}_1),\\
&d_{m,1}^D=(\mathbf{K}_{m}^1\bm{\alpha}_1)^\top(\mathbf{K}_{m}^1\bm{\alpha}_1).
\end{split}
\]

To simplify the denotation, in the proof, $d_{m,p}$ expresses the kernel weight $d_{m,p}^t$ of the $t^{\text{th}}$ iteration of $\bm{\alpha}_p$ and $\mathbf{d}_p$.

\subsection{Proof of First Conclusion}
According to Eq.~(\ref{eq:alpha-emkcf}), we set $\bm{\alpha}_p=\mathbf{D}^{-1}_p\mathbf{N}_p\mathbf{y}_c$, where
\[
\mathbf{D}_p=\displaystyle\sum_{j=1}^p\sum_{m=1}^M\beta^j_{m}\left((d_{m,p}\mathbf{K}_{m}^j)^2+\lambda d_{m,p}\mathbf{K}_{m}^j\right)
\]
and
\[
\mathbf{N}_p=\displaystyle\sum_{j=1}^p\sum_{m=1}^M\beta_{m}^j d_{m,p}\mathbf{K}_{m}^j.
\]
It is clear that both $\mathbf{D}_p$ and $\mathbf{N}_p$ are positive definite because $\beta^j_{m}>0$, $d_{m,p}>0$, $\lambda>0$, and $\mathbf{K}_{m}^j$ is positive definite. Because $\mathbf{K}_m^j$ is circulant Gram matrix, we have $\mathbf{K}_m^j=\mathbf{U}\mathbf{\Sigma}_m^j\mathbf{U}^H$, where $\mathbf{U}=\frac{1}{\sqrt{l}}\mathbf{F}_l^{-1}$ and $\mathbf{F}_l$ is the 1-D discrete Fourier transform matrix~\cite{davidp94}.
Because the linear combination of circulant matrices is also circulant, we have
\[
\mathbf{D}_p=\mathbf{U}\left(\displaystyle\sum_{j=1}^p\sum_{m=1}^M\beta^j_{m}\left(d_{m,p}^2(\mathbf{\Sigma}_m^j)^2+\lambda d_{m,p}\mathbf{\Sigma}_m^j\right)\right)\mathbf{U}^H
\]
and
\[
\mathbf{N}_p=\mathbf{U}\left(\displaystyle\sum_{j=1}^p\sum_{m=1}^M\beta^j_{m}d_{m,p}\mathbf{\Sigma}_m^j\right)\mathbf{U}^H.
\]
Let $\bm{\Sigma}_m^j=\mathbf{diag}\left(\sigma_{m,1}^j,\ldots,\sigma_{m,l}^j\right)$, $\sigma_{m,n}^j>0$, $n=1,\dots,l$. Then the $n^{\text{th}}$ eigenvalue of $\mathbf{D}_p^{-1}\mathbf{N}_p$ is
\[
\begin{split}
\sigma_{\bm{\alpha}_p,n} & \equiv\displaystyle\frac{\sum_{j=1}^p\sum_{m=1}^M\beta^j_{m}d_{m,p}\mathbf{\sigma}_{m,n}^j}{\sum_{j=1}^p\sum_{m=1}^M\beta^j_{m}d_{m,p}\mathbf{\sigma}_{m,n}^j(d_{m,p}
\sigma_{m,n}^j+\lambda)}\\
& =(\lambda+b_{n})^{-1},
\end{split}
\]
where
\[
b_{n}=\displaystyle\frac{\sum_{j=1}^p\sum_{m=1}^M\beta^j_{m}d_{m,p}^2(\sigma_{m,n}^j)^2}{\sum_{j=1}^p\sum_{m=1}^M
\beta^j_{m}d_{m,p}\sigma_{m,n}^j}.
\]
It is clear that $b_{n}>0$.

According to Eq.~(\ref{eq:d-evaluation}), we also have
\[
\begin{split}
   d_{m,p}^N & = \sum_{j=1}^{p}\beta_{m}^j(\mathbf{K}_{m}^j\bm{\alpha}_p)^\top(2\mathbf{y}_c-\lambda\bm{\alpha}_p)\\
     & = \mathbf{y}_c^{\top}\sum_{j=1}^{p}\beta_{m}^j\mathbf{N}_p\mathbf{D}^{-1}_p\mathbf{K}_m^j(2\mathbf{I}-\lambda\mathbf{D}^{-1}_p\mathbf{N}_p)\mathbf{y}_c \\
     & = \mathbf{y}_c^{\top}\mathbf{D}_{m,p}^N\mathbf{y}_c,
\end{split}
\]
where $\mathbf{D}_{m,p}^N=\mathbf{N}_p\mathbf{D}^{-1}_p\sum_{j=1}^{p}\beta_{m}^j\mathbf{K}_m^j(2\mathbf{I}-\lambda\mathbf{D}^{-1}_p\mathbf{N}_p)$, and its $n^{\text{th}}$ eigenvalue is
\[
\sigma_{m,p,n}^N=\sigma_{\bm{\alpha}_p,n}(2-\lambda\sigma_{\bm{\alpha}_p,n})\sum_{j=1}^{p}\beta_{m}^j\sigma_{m,n}^j.
\]
$\because$ $\lambda\sigma_{\bm{\alpha}_p,n}=\lambda(\lambda+b_{n})^{-1}<1$,
$\therefore$ $2-\lambda\sigma_{\bm{\alpha}_p,n}>1$, $\therefore$ $\sigma_{m,p,n}^N>0$, $n=1,\ldots,l$, $\therefore$ $\mathbf{D}_{m,p}^N$ is positive definite, and $d_{m,p}^N>0$. It is obvious that $d_{m,p}^D>0$. Consequently,
\[
d_{m,p}^{t+1}=\frac{d_{m,p}^N}{d_{m,p}^D}>0,
\]
where $m=1,\ldots,M$.

\vspace{3mm}
\noindent Q.E.D.

\subsection{Proof of Second Conclusion}
According to Eq.~(\ref{eq:alpha-emkcf}), we have
\[
\begin{split}
   d_{m,p}^D & =2\sum_{j=1}^{p}\beta_{m}^j(\mathbf{K}_{m}^j\bm{\alpha}_p)^\top(\mathbf{K}_{m}^j\bm{\alpha}_p) \\
     & = \mathbf{y}_c^{\top}2\sum_{j=1}^{p}\beta_{m}^j\mathbf{N}_p\mathbf{D}^{-1}_p(\mathbf{K}_m^j)^2\mathbf{D}^{-1}_p\mathbf{N}_p\mathbf{y}_c \\
     & = \mathbf{y}_c^{\top}\mathbf{D}_{m,p}^D\mathbf{y}_c，
\end{split}
\]
where $\mathbf{D}_{m,p}^D=2\mathbf{N}_p\mathbf{D}^{-1}_p\sum_{j=1}^{p}\beta_{m}^j(\mathbf{K}_m^j)^2\mathbf{D}^{-1}_p\mathbf{N}_p$, and its $n^{\text{th}}$ eigenvalue is
\[
\sigma_{m,p,n}^D=2\sigma_{\bm{\alpha}_p,n}^2\sum_{j=1}^{p}\beta_{m}^j(\sigma_{m,n}^j)^2.
\]
Then, according to Eq.~(\ref{eq:d-evaluation}),
\[
d_{m,p}^{t+1}=\frac{d_{m,p}^N}{d_{m,p}^D}=\frac{\mathbf{y}_c^{\top}\mathbf{D}_{m,p}^N\mathbf{y}_c}{\mathbf{y}_c^{\top}\mathbf{D}_{m,p}^D\mathbf{y}_c}=
\frac{\mathbf{y}_c^{\top}\mathbf{U}\bm{\Sigma}_{m,p}^N\mathbf{U}^H\mathbf{y}_c}{\mathbf{y}_c^{\top}\mathbf{U}\bm{\Sigma}_{m,p}^D\mathbf{U}^H\mathbf{y}_c}.
\]
Let $\mathbf{U}^H\mathbf{y}_c=(y_{u,1},\ldots,y_{u,l})$, $c_n^N=\sum_{j=1}^{p}\beta_{m}^j\sigma_{m,n}^j$, and $c_n^D=\sum_{j=1}^{p}\beta_{m}^j(\sigma_{m,n}^j)^2$. Then
\[
d_{m,p}^{t+1}=\frac{\sum_{n=1}^{l}y_{u,n}^2 c_n^N\sigma_{\bm{\alpha}_p,n}(2-\lambda\sigma_{\bm{\alpha}_p,n})}{2\sum_{n=1}^{l}y_{u,n}^2 c_n^D\sigma_{\bm{\alpha}_p,n}^2}.
\]
Let $c_{\max}^N=\max_n c_n^N$, $c_{\min}^N=\min_n c_n^N$, $c_{\max}^D=\max_n c_n^D$, $c_{\min}^D=\min_n c_n^D$, $y_{\max}=\max_n y_{u,n}$, $y_{\min}=\min_n y_{u,n}$,
\[
c_l=\frac{y_{\min}^2c_{\min}^N}{y_{\max}^2c_{\max}^N},\;\;c_u=\frac{y_{\max}^2c_{\max}^N}{y_{\min}^2c_{\min}^N},
\]
and
\[
\sigma_r=\frac{\sum_{n=1}^{l}\sigma_{\bm{\alpha}_p,n}(2-\lambda\sigma_{\bm{\alpha}_p,n})}{2\sum_{n=1}^{l}\sigma_{\bm{\alpha}_p,n}^2}.
\]
Then
\[
c_l\cdot\sigma_r<d_{m,p}^{t+1}<c_u\cdot\sigma_r.
\]
Furthermore,
\[
\begin{split}
\sigma_r & =\frac{\sum_{n=1}^{l}(\lambda+b_{n})^{-1}(2(\lambda+b_{n})-\lambda)(\lambda+b_{n})^{-1}}{2\sum_{n=1}^{l}(\lambda+b_{n})^{-2}}\\
         & =\frac{\sum_{n=1}^{l}(\lambda+b_{n})^{-2}(\lambda+2b_{n})}{2\sum_{n=1}^{l}(\lambda+b_{n})^{-2}}\\
         & =\frac{1}{2}\lambda+\frac{\sum_{n=1}^{l}b_{n}(\lambda+b_{n})^{-2}}{\sum_{n=1}^{l}(\lambda+b_{n})^{-2}}.
\end{split}
\]
Let $\sigma_{m,\max}^j=\max_{n}\sigma_{m,n}^j$, $\sigma_{m,\min}=\min_{n}\sigma_{m,n}^j$,
\[
\begin{split}
& b^{\max}=\frac{\sum_{j=1}^p\sum_{m=1}^M\beta^j_{m}d_{m,p}^2(\sigma_{m,\max}^j)^2}{\sum_{j=1}^p\sum_{m=1}^M,
\beta^j_{m}d_{m,p}\sigma_{m,\min}^j},\\
&b^{\min}=\frac{\sum_{j=1}^p\sum_{m=1}^M\beta^j_{m}d_{m,p}^2(\sigma_{m,\min}^j)^2}{\sum_{j=1}^p\sum_{m=1}^M,
\beta^j_{m}d_{m,p}\sigma_{m,\max}^j}.
\end{split}
\]
Then, $b^{\min}\leq b_n\leq b^{\max}$, and
\[
\frac{1}{2}\lambda+b^{\min}\leq\sigma_r\leq\frac{1}{2}\lambda+b^{\max},
\]
\[
\frac{c_l}{2}\lambda+c_l\cdot b^{\min}<d_{m,p}^{t+1}<\frac{c_u}{2}\lambda+c_u\cdot b^{\max}.
\]
where $m=1,\ldots,M$.

\vspace{3mm}
\noindent Q.E.D.

{\small
\bibliographystyle{ieee}
\bibliography{bibjournal}
}